\documentclass[fleqn,10pt]{wlscirep}
\usepackage{longtable}
\usepackage{array} 
\usepackage{xcolor,amsmath}
\usepackage{float}
\usepackage{amsfonts}
\usepackage{dsfont}
\usepackage{yfonts}
\usepackage{url}
\usepackage{subfigure}
\usepackage{comment}
\usepackage[mathcal]{eucal}
\usepackage{multirow}
\usepackage{makecell}
\usepackage{bbding}
\usepackage{algorithm}
\usepackage[noend]{algpseudocode}

\usepackage{epstopdf}
\usepackage{graphicx}
\usepackage{bbm}

\usepackage{soul}
\soulregister\cite7 
\soulregister\citep7 
\soulregister\citet7 
\soulregister\ref7 
\soulregister\pageref7 
\soulregister\it7 

\usepackage{wrapfig}
\usepackage{booktabs}
\usepackage{graphicx}%
\usepackage{multirow}%
\usepackage{amsmath,amssymb,amsfonts}%
\usepackage{amsthm}%
\usepackage{mathrsfs}%
\usepackage[title]{appendix}%
\usepackage{xcolor}%
\usepackage{textcomp}%
\usepackage{manyfoot}%
\usepackage{booktabs}%
\usepackage{array}
\usepackage{algorithm}%
\usepackage{algorithmicx}%
\usepackage{algpseudocode}%
\usepackage{listings}%

\usepackage{multirow}
\usepackage[normalem]{ulem}
\usepackage{algorithm}

\usepackage{subfigure}  
\usepackage{multirow} 
\usepackage{soul}

\usepackage{booktabs}      
\usepackage{arydshln}      
\usepackage{ulem}          

\renewcommand{\ul}[1]{\underline{#1}}
\usepackage{xcolor}

\usepackage{amsmath}
\usepackage{booktabs}
\usepackage{amssymb}  
\usepackage{newfloat}

\title{\LARGE NGTM: Substructure-based Neural Graph Topic Model for Interpretable Graph Generation}

\author[1]{Yuanxin Zhuang}
\author[2]{Dazhong Shen}
\author[1*]{Ying Sun}

\affil[1]{The Thrust of Artificial Intelligence, The Hong Kong University of Science and Technology (Guangzhou), Guangzhou, China.}
\affil[2]{The College of Computer Science and Technology, The Nanjing University of Aeronautics and Astronautics, Nanjing, China}
\affil[*]{The corresponding authors (yings@hkust-gz.edu.cn).}

\begin{abstract}
Graph generation plays a pivotal role across numerous domains, including molecular design and knowledge graph construction. Although existing methods achieve considerable success in generating realistic graphs, their interpretability remains limited, often obscuring the rationale behind structural decisions. To address this challenge, we propose the Neural Graph Topic Model (NGTM), a novel generative framework inspired by topic modeling in natural language processing. NGTM represents graphs as mixtures of latent topics, each defining a distribution over semantically meaningful substructures, which facilitates explicit interpretability at both local and global scales. The generation process transparently integrates these topic distributions with a global structural variable, enabling clear semantic tracing of each generated graph. Experiments demonstrate that NGTM achieves competitive generation quality while uniquely enabling fine-grained control and interpretability, allowing users to tune structural features or induce biological properties through topic-level adjustments.
\end{abstract}

\begin{document}
\maketitle

\section{Introduction}\label{sec1}

Graph generation is a fundamental task with diverse applications—including molecular design~\cite{tian2025leveraging}, knowledge graph construction~\cite{wu2025construction}, and relational data modeling~\cite{rigoni2025rgcvae}. While generating valid and accurate graphs is essential, understanding the underlying generative process and explaining structure formation is equally crucial~\cite{kumar2025explainable}. This significance is particularly evident in high-stakes applications like drug discovery~\cite{zheng2024application}, where generation process interpretability builds trust, enables rational design, and supports downstream scientific analysis~\cite{doshi2017towards, holzinger2019causability, tjoa2020survey}.

Despite its importance, interpretability remains a major challenge in existing graph generation models~\cite{guo2022systematic, zhang2023survey, liu2023generative}. Sequential methods, such as the auto-regressive model Graph Generative Pre-trained Transformer (G2PT)~\cite{chen2025graph} and the reinforcement learning-based model ExSelfRL~\cite{wang2025exselfrl}, generate graphs step-by-step by incrementally adding nodes and edges. While their procedures are explicit, the underlying decisions depend heavily on hidden states, making semantic interpretation difficult.
In contrast, models based on VAEs like DAVA~\cite{hou2024dag} and GANs like ConfGAN~\cite{xu2025generation} generate entire graphs from latent representations in a single forward pass, while diffusion-based models like ConStruct~\cite{madeira2024generative} iteratively refine graphs from random noise. Although these methods achieve strong performance and capture complex distributions, their latent spaces are often opaque and unaligned with interpretable graph components. As a result, they struggle to provide insights into the reasoning behind generated graphs.

Graphs are often composed of key substructures that define their topology and functionality~\cite{kengkanna2024enhancing}. For example, molecular graphs contain recurring building blocks like functional groups or rings, which are essential to their structural diversity and chemical behavior~\cite{jin2018junction}. Explicitly modeling these substructures improves the quality and diversity of generated graphs while uncovering hidden patterns that reveal new domain knowledge~\cite{kong2022principal, jin2020multi}. Additionally, this approach enhances generation transparency, enables precise control over substructure inclusion, and ensures generated graphs align with real-world characteristics.

However, achieving interpretable graph generation remains challenging. Post hoc methods analyze latent features or learned representations after generation to identify meaningful patterns in graphs~\cite{stoehr2019disentangling, guo2020interpretable}. While useful, these approaches do not influence the generative process itself, leaving the reasoning behind structural decisions untraceable. Some generative models construct molecular graphs from substructures~\cite{jin2018junction, jin2020multi, kong2022principal}. Although effective in improving validity and modularity, these methods have two key limitations: (1) they often struggle to automatically discover and organize previously unseen substructures into coherent semantic units; and (2) they often lack a unified, transparent narrative of the generation process, making it difficult to trace the relationships and contributions of individual substructures to the overall graph. Moreover, such reliance on predefined or pre-extracted substructures can constrain generative flexibility and hinder the discovery of novel structures.

To address these challenges, we propose the Neural Graph Topic Model (NGTM), a novel interpretable generative framework inspired by topic modeling techniques in natural language processing~\cite{blei2003latent}. NGTM models graphs as mixtures of latent topics, with each topic corresponding to a distinct distribution over interpretable graph substructures (e.g., rings, motifs, functional groups in molecular graphs). By explicitly extracting reusable structural units and organizing them into semantically coherent topics, NGTM achieves clear semantic organization in graph generation.
Based on the topics, the generation process in NGTM is transparent and traceable, comprising three clear steps: (1) sampling a topic mixture to define the semantic profile of the target graph; (2) sampling relevant substructures based on these sampled topics; and (3) assembling these substructures under learned structural constraints to form a coherent graph, thus integrating both local interpretability (via explicit substructures) and global interpretability (via topic mixtures) in a unified and transparent way.
Our experiments reveal that NGTM not only achieves competitive generation quality but, more importantly, pioneers a new level of interpretability and controllability in graph generation. The learned topics align robustly with meaningful structural and functional classes (e.g., carcinogenicity), providing a clear semantic basis. By adjusting these topics, NGTM enables continuous and fine-grained control over key graph properties, allowing the generation process to be precisely steered toward desired outcomes.

\section{Results}

\begin{figure*}
  \centering
  \includegraphics[width=1\textwidth]{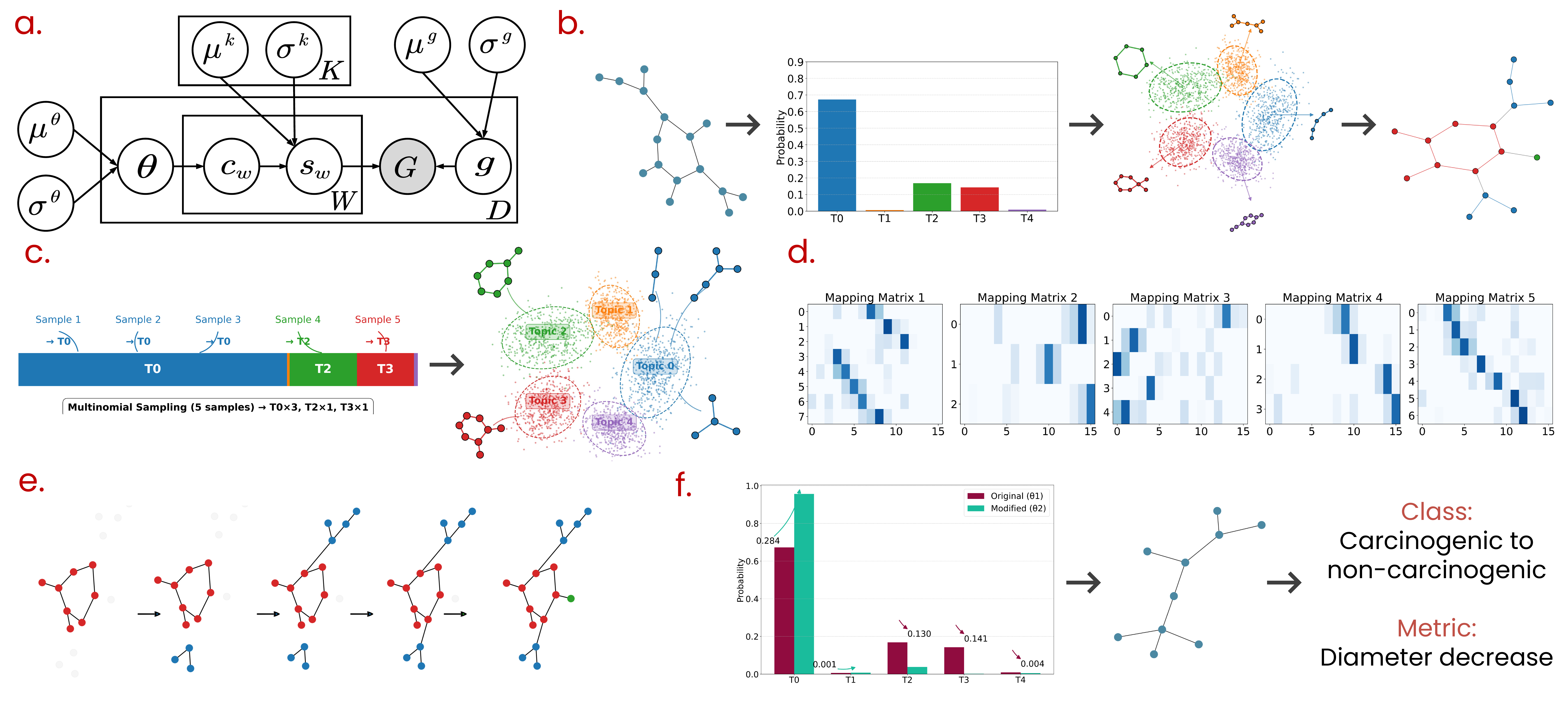}
  \vspace{-15pt}
  \caption{Figure 1. Overview of the NGTM framework for interpretable graph generation.
  (a) The probabilistic graphical model of NGTM. (b) Training phase: NGTM infers topic mixtures from real graphs and discovers semantically meaningful substructure topics, enabling reconstruction through interpretable latent factors.
  (c–e) Generation phase: (c) Multinomial sampling from $\theta$ assigns each substructure to a topic, and the corresponding latent vectors are decoded into interpretable substructures. (d) Mapping matrices determine how substructures are softly aligned and integrated into the growing graph. (e) Visualization of the sequential assembly process guided by global structure vectors.
  (f) Example of controllable generation: adjusting topic proportions modifies graph semantics.}
  \label{fig:ngtm_model_generation}
  \vspace{-10pt}
\end{figure*}

\subsection{Overview of the NGTM Graph Generation Process}

Graphs can be viewed as compositions of structural motifs or substructures, each originating from a semantically meaningful latent topic. For example, molecular graphs typically contain interpretable substructures like rings, chains, and functional groups that frequently co-occur. To explicitly capture this structural diversity, we propose NGTM, a novel and interpretable generative approach that models graph structures as assemblies of substructures sampled from a set of latent topics. Specifically, we assume there are $K$ latent topics, each topic defines a Gaussian distribution over latent embeddings of substructures, denoted as $\Phi = {\mathcal{N}(\mu^k, \sigma^k)}_{k=1}^{K}$. These distributions are parameterized and learned through a Conditional Variational Autoencoder (CVAE)~\cite{sohn2015learning}, with topics serving as conditioning variables during training. Such a design enables the generation of diverse and potentially novel substructures without relying on a predefined substructure vocabulary. Latent vectors sampled from these topic-specific distributions are decoded into adjacency matrices representing meaningful subgraphs.
To govern the semantic composition, we introduce a latent variable drawn from $\mathcal{N}(\mu^\theta, \sigma^\theta)$, transformed via softmax into a topic proportion vector $\theta$. This vector directs the topic selection for substructure generation, ensuring semantic coherence in the resulting graph.
Additionally, we introduce a global structural variable $g \sim \mathcal{N}(\mu^g, \sigma^g)$ to capture high-level topological properties such as density, connectivity, and overall layout. This global guidance ensures that the final assembled graph is coherent, realistic, and structurally sound.
The NGTM generative framework is outlined in Figure~\ref{fig:ngtm_model_generation}, where Figure~\ref{fig:ngtm_model_generation}(a) shows the probabilistic graphical representation detailing the relationships between latent variables involved in the generation process.
Figure~\ref{fig:ngtm_model_generation}(b) illustrates the training phase, during which NGTM learns topic mixtures from observed graphs and organizes substructures into semantically coherent topics to support interpretable reconstruction.
Figures~\ref{fig:ngtm_model_generation}(c) to (e) depict the generation process: (c) shows how topic assignments are first drawn from the sampled topic proportion vector $\theta$, and corresponding substructures are generated by sampling latent vectors from the assigned topic-specific distributions.
(d) depicts the construction of soft mapping matrices for each substructure under the guidance of the global structural vector $g$, determining how substructures are integrated into the current graph. (e) visualizes the step-by-step assembly of the full graph by sequentially merging substructures according to the learned mappings and global structural constraints.
Figure~\ref{fig:ngtm_model_generation}(f) demonstrates controllable generation, where adjusting topic weights results in interpretable changes to graph-level properties, such as reducing diameter or altering predicted class labels. The full generative process of NGTM can be summarized in the following three stages:

1. Global and semantic initialization:

\quad \quad a. Sample latent semantic vector $z^\theta \sim \mathcal{N}(\mu^{\theta}, \sigma^{\theta})$ and derive topic proportion vector $\theta = \text{softmax}(z^\theta)$.

\quad \quad b. Sample global structural guidance vector $g \sim \mathcal{N}(\mu^{g}, \sigma^{g})$.

2. Substructure generation loop (for $w = 1, \ldots, W$):

\quad \quad a. Select topic $c_w \sim \text{Multinomial}(\theta)$.
   
\quad \quad b. Generate latent substructure vector $z_w \sim \mathcal{N}(\mu^{k=c_w}, \sigma^{k=c_w})$ and decode substructure $s_w$.

3. Final assembly: Combine substructures and global guidance: $G = f(s_1, s_2, \dots, s_W, g)$.

\subsection{Experimental Setup}

\subsubsection{Baselines} 
We compare our approach with several state-of-the-art graph generation models, covering a diverse set of generative paradigms:
\textbf{(1) GraphVAE}~\cite{GraphVAE}: A variational autoencoder (VAE)-based model that generates graphs in a one-shot manner using latent embeddings.
\textbf{(2) GraphVAE-MM}~\cite{GraphVAE–MM}: An enhanced version of GraphVAE trained with a micro-macro objective that balances local and global structural fidelity.
\textbf{(3) GraphRNN}~\cite{GraphRNN}: An autoregressive model that generates graphs sequentially, modeling the distribution over adjacency vectors using RNNs.
\textbf{(4) GRAN}~\cite{GRAN}: Graph Recurrent Attention Networks generate graphs in blocks of nodes and their edges using attention mechanisms.
\textbf{(5) BiGG}~\cite{BiGG}: A scalable autoregressive model for generating large and sparse graphs using graph partitioning and latent codes.
\textbf{(6) DiGress}~\cite{vignac2022digress}: A discrete diffusion model that treats graph generation as a denoising classification task over discrete node and edge types, incorporating structural features to improve expressiveness.
\textbf{(7) G2PT}~\cite{chen2025graph}: A pre-trained autoregressive graph generation framework that represents graphs as sequences and leverages transformer architectures with task-agnostic fine-tuning capabilities.
\textbf{(8) ConStruct}~\cite{madeira2024generative}: A constraint-aware diffusion model that embeds structural constraints such as planarity or acyclicity directly into the generation process through projection operators and noise design.
\textbf{(9) NGTM Variants:} We further compare against ablated versions of our NGTM model: (1) \textbf{NGTM$_\text{w/oGE}$:} This variant removes the global encoder, assessing the importance of global structural guidance. (2) \textbf{NGTM$_\text{Parallel}$:} In this version, the Substructure Assembly Module performs simultaneous aggregation of all substructures rather than sequential composition.

\subsubsection{Datasets}
We utilize one synthetic dataset and three real-world datasets:
\textbf{(1) Lobster}~\cite{golomb1996polyominoes}: This dataset includes 100 synthetic graphs with $10 \leq |V| \leq 100$. The average number of nodes is 52. These graphs are trees where each node is at most 2 hops away from a backbone path.
\textbf{(2) MUTAG}~\cite{debnath1991structure}: MUTAG is a dataset with $10 \leq |V| \leq 28$ comprising 188 mutagenic aromatic and heteroaromatic nitro compounds. The average number of nodes is 18.
\textbf{(3) PTC}~\cite{toivonen2003statistical, xu2018powerful}: PTC is a dataset of 344 chemical compounds with $4 \leq |V| \leq 103$ that report the carcinogenicity of male and female rats. The average number of nodes is 26.
\textbf{(4) Ogbg-molbbbp}~\cite{hu2020open}: This dataset consists of 2039 real-world molecular graphs with $2 \leq |V| \leq 132$. The average number of nodes is 23.
\textbf{Train/Test Split.} We follow the same protocol as~\cite{GraphRNN, BiGG, GraphVAE–MM} and create random 80\% and 20\% splits of the graphs for training and testing, respectively. Additionally, 20\% of the training data in each split is used as the validation set.

\subsubsection{Implementation Details}
Training is performed using the Adam optimizer with a learning rate of 0.0003. The model is trained for up to 20,000 epochs, and the version with the best validation performance is selected for testing.
We fix the number of topics \(K = 10\), the number of sampled substructures \(W = 30\), and set the substructure size \(n\) to the average number of nodes in each dataset. We fix the number of topics \(K = 10\), the number of sampled substructures \(W = 30\), and set the substructure size \(n\) to the average number of nodes in each dataset. Although these hyperparameters may not yield the best possible reconstruction accuracy, they provide a balance between interpretability and generation quality in our experiments.
The NGTM framework is composed of modular components: a topic encoder, structure encoder, and global encoder—each implemented with four GCN layers, followed by Layer Normalization and average pooling. The structure decoder consists of three linear layers. For substructure assembly, we employ three MultiheadAttention modules to compute attention between the substructure, the current graph, and the global structure vector. The mapping network includes three linear layers with LayerNorm. All hidden dimensions are set to 256.

\subsubsection{Evaluation Metrics}
We evaluate the effectiveness of NGTM through both qualitative and quantitative metrics to comprehensively assess the diversity and realism of the generated graphs.
\textbf{Qualitative Evaluation:} This involves visually inspecting the generated graphs and comparing them with real samples to assess structural plausibility and visual similarity. Detailed visual comparisons for each model across various datasets are provided in the Appendix (see Appendix~\ref{appendix_visualization}).
\textbf{Quantitative Evaluation:} We use two types of metrics to measure the distributional distance between generated and test graphs:
(1) \textbf{GNN-based Metrics:} 
These metrics utilize a task-agnostic Graph Neural Network (Random-GNN)~\cite{thompson2022evaluation} to extract graph representations. We evaluate the discrepancies using F1 PR, which measures the diversity of the generated graphs, and MMD RBF, which assesses their realism based on structural and semantic alignment.
(2) \textbf{Statistics-based Metrics:} 
This approach directly compares structural statistics such as degree distributions, orbit counts, clustering coefficients, spectral features, and graph diameter~\cite{GRAN, GraphVAE–MM}. Following O’Bray et al.~\cite{OBray_Horn_Rieck_Borgwardt_2021}, we also report ideal scores obtained from a 50/50 data split, which serve as a lower bound for distributional distance.
These combined metrics provide a comprehensive view of the model’s performance.

\begin{table*}[t!]
\centering
\caption{Comparison of NGTM and baselines. Best results are \textbf{bold}, second best are \underline{underlined}.}
\vspace{-10pt}
\resizebox{1\textwidth}{!}{
\begin{tabular}{cccccccc|ccccccc}
\specialrule{0em}{1.5pt}{1.5pt}
\toprule
\midrule[1pt]
\multirow{2}{*}{\textbf{Method}} & \multicolumn{7}{c|}{\textbf{MUTAG}}                                                                              & \multicolumn{7}{c}{\textbf{Lobster}}                                                                               \\
                                 & Deg. $\downarrow$          & Clus. $\downarrow$         & Orbit $\downarrow$          & Spect $\downarrow$          & Diam. $\downarrow$          & MMD $\downarrow$       & F1 $\uparrow$          & Deg. $\downarrow$          & Clus. $\downarrow$         & Orbit $\downarrow$          & Spect $\downarrow$          & Diam. $\downarrow$          & MMD $\downarrow$       & F1 $\uparrow$          \\ \hline
50/50 split                              & 3e-4              & 0                  & 2e-5               & 0.007              & 0.002              & 0.027            & 97.96          & 8e-4              & 0                  & 0.003              & 0.004              & 0.023              & 0.04             & 98.99          \\
GraphRNN                                 & 0.007             & 0.266              & 0.001              & 0.070              & 0.728              & 0.832            & 52.81          & 0.004             & 0                  & 0.039              & 0.044              & 0.376              & 0.855            & 62.33          \\
GRAN                                     & \ul{5e-4}        & \ul{0.011}        & 0.005              & 0.059              & 0.645              & 0.276            & 90.87          & 0.006             & 0.475              & 0.412              & 0.039              & 0.502              & 0.221            & 45.39          \\
BiGG                                     & 0.006             & \ul{0}            & 0.003              & 0.043              & 0.308              & 0.521            & 97.93          & 5e-4              & 0                  & \textbf{0.001}     & 0.017              & \ul{0.009}        & 0.127            & 95.19          \\
GraphVAE                                 & 0.006             & 0.122              & 0.007              & 0.024              & 0.051              & 0.109            & 69.79          & 0.094             & 0.866              & 0.405              & 0.071              & 0.137              & 0.415            & 69.54          \\
GraphVAE-MM                              & 0.001             & 0                  & 5e-4               & \ul{0.017}        & 0.021              & 0.061            & 89.58          & 4e-4              & 0                  & 0.008              & 0.019              & 0.175              & 0.136            & 100            \\
DiGress                                  & 0.002             & 0.015              & 7e-4               & 0.020              & 0.060              & 0.085            & 95.23          & 0.001             & 0.045              & 0.006              & 0.030              & 0.038              & 0.148            & 90.44          \\
G2PT                                     & 0.004             & 0.052              & 0.003              & 0.029              & 0.087              & 0.132            & 97.66          & 0.003             & 0.076              & 0.005              & 0.034              & 0.051              & 0.193            & 95.12          \\
ConStruct                                & 0.001             & 0                  & 0.001              & 0.018              & \ul{0.020}        & \ul{0.059}      & 94.12          & \textbf{2e-4}     & 0                  & 0.004              & \ul{0.015}        & 0.010              & \textbf{0.097}   & 91.36          \\ \hdashline
NGTM$_\text{w/oGE}$    & 0.005             & 0.041              & 9e-4               & 0.023              & 0.079              & 0.082            & 96.73          & 0.002             & 0.071              & 0.009              & 0.035              & 0.027              & 0.139            & \ul{98.75}    \\
NGTM$_\text{Parallel}$ & 0.006             & 0                  & \ul{4e-4}         & 0.018              & 0.050              & 0.068            & \ul{98.67}    & 0.003             & \ul{0.021}        & 0.012              & 0.016              & 0.017              & 0.159            & 100            \\
NGTM                                     & \textbf{4e-4}     & \textbf{0}         & \textbf{3e-4}      & \textbf{0.015}     & \textbf{0.015}     & \textbf{0.053}   & \textbf{98.72} & \ul{3e-4}        & \textbf{0}         & \ul{0.003}        & \textbf{0.012}     & \textbf{0.007}     & \ul{0.102}      & \textbf{100}   \\ \hline
& \multicolumn{7}{c|}{\textbf{PTC}}                                                                                  & \multicolumn{7}{c}{\textbf{Ogbg-molbbbp}}                                                                           \\ \hline
50/50 split                              & 5e-5              & 3e-4               & 7e-5               & 0.003              & 0.008              & 0.022            & 96.73          & 5e-4              & 4e-5               & 8e-5               & 8e-4               & 7e-4               & 0.001            & 97.98          \\
GraphRNN                                 & 0.007             & 0.004              & 0.006              & 0.071              & 0.417              & 0.816            & 33.15          & 0.003             & 0.001              & 9e-4               & 0.126              & 0.553              & 1.361            & 94.73          \\
GRAN                                     & 0.020             & 0.115              & 0.009              & 0.038              & 0.179              & 0.164            & 78.84          & 0.006             & 0.376              & 0.011              & 0.044              & 0.359              & 0.354            & 93.96          \\
BiGG                                     & \textbf{3e-4}     & 0.004              & \textbf{9e-5}      & \ul{0.019}        & \ul{0.027}        & 0.056            & \ul{95.89}    & 0.002             & \ul{0.001}        & 9e-4               & 0.009              & 0.031              & 0.059            & 96.22          \\
GraphVAE                                 & 0.223             & 0.729              & 0.537              & 0.035              & 0.204              & 0.613            & 42.31          & 0.031             & 0.562              & 0.048              & 0.019              & 0.061              & 0.313            & 57.38          \\
GraphVAE-MM                              & 0.033             & \ul{5e-4}         & \ul{0.002}        & 0.022              & 0.038              & 0.055            & 82.39          & 0.002             & 0.007              & 9e-4               & \textbf{0.004}     & 0.031              & 0.047            & 92.96          \\
DiGress                                  & 0.012             & 0.008              & 0.007              & 0.026              & 0.098              & 0.113            & 89.85          & 0.003             & 0.005              & 0.001              & 0.010              & 0.042              & 0.073            & 93.67          \\
G2PT                                     & 0.022             & 0.016              & 0.010              & 0.033              & 0.079              & 0.104            & 93.70          & \ul{0.002}       & 0.008              & \ul{6e-4}         & 0.013              & 0.048              & 0.092            & \textbf{96.85} \\
ConStruct                                & 0.005             & 0.001              & 0.003              & 0.021              & 0.032              & \ul{0.051}      & 94.73          & 0.001             & \textbf{5e-4}      & 0.001              & \ul{0.006}        & \ul{0.030}        & \ul{0.052}      & 92.18          \\ \hdashline
NGTM$_\text{w/oGE}$    & 0.029             & 0.008              & 0.027              & 0.053              & 0.142              & 0.178            & 90.82          & 0.004             & 0.007              & 9e-4               & 0.033              & 0.097              & 0.271            & 85.39          \\
NGTM$_\text{Parallel}$ & 0.056             & 7e-4               & 0.014              & 0.067              & 0.398              & 0.367            & 83.71          & 0.001             & 0.009              & 7e-4               & 0.026              & 0.245              & 0.289            & 79.24          \\
NGTM                                     & \ul{0.002}       & \textbf{2e-4}      & 0.003              & \textbf{0.017}     & \textbf{0.022}     & \textbf{0.046}   & \textbf{96.03} & \textbf{0.001}    & 0.004              & \textbf{5e-4}      & 0.008              & \textbf{0.029}     & \textbf{0.032}   & \ul{96.38}    \\ \midrule[1pt] 
\specialrule{0em}{1.5pt}{1.5pt}
\bottomrule
\end{tabular}}
\label{tab:mainall}
\end{table*}

\subsection{Comprehensive Evaluation of Generation Performance}

To comprehensively evaluate the generation capabilities of NGTM, we conduct three sets of experiments. First, in Generation Quality Evaluation, we compare NGTM against state-of-the-art baselines and analyze ablation variants to understand the contribution of each module. Second, we perform Parameter Sensitivity analysis to investigate the impact of key hyperparameters on performance. Finally, we carry out Evaluation on Capturing Class-wise Structural Differences to assess NGTM's ability to replicate meaningful structural variations across semantic classes.

\subsubsection{Generation Quality Evaluation}
We conducted experiments comparing NGTM with several graph generation models across four representative datasets, spanning synthetic, chemical, and bioinformatics domains. As shown in Table~\ref{tab:mainall}, NGTM consistently achieves results that rival or outperform the strongest baselines, despite explicitly emphasizing interpretability—a quality often seen as coming at the cost of generation fidelity.

Notably, our evaluations show that NGTM is capable of robustly capturing both local structural motifs and global topological characteristics, essential for realistic graph synthesis. This is particularly significant in chemically relevant scenarios such as MUTAG, PTC, and Ogbg-molbbbp datasets, where generating accurate molecular structures directly impacts real-world outcomes, including drug design and toxicity prediction. NGTM-generated graphs exhibit clear and consistent chemical substructures that align well with known domain knowledge, demonstrating its strong practical potential in scientific contexts, as further illustrated by the examples in Appendix Figure~\ref{fig:Visualizations}.

Further analysis through ablation studies highlights crucial insights into NGTM’s design. Removing the global structural guidance module (\textit{NGTM}\textsubscript{w/oGE}) markedly degraded the coherence and realism of generated graphs. This observation clearly illustrates that maintaining high-level global structure constraints significantly enhances the model’s ability to produce realistic and semantically coherent graphs. Similarly, replacing sequential substructure assembly with parallel integration (\textit{NGTM}\textsubscript{Parallel}) adversely affected structural consistency, suggesting that our sequential assembling of semantic components has been critical for preserving complex structural dependencies.

\begin{figure}[]
  \centering
  \subfigure[Substructure Count ($W$)]{
      \includegraphics[width=0.3\textwidth]{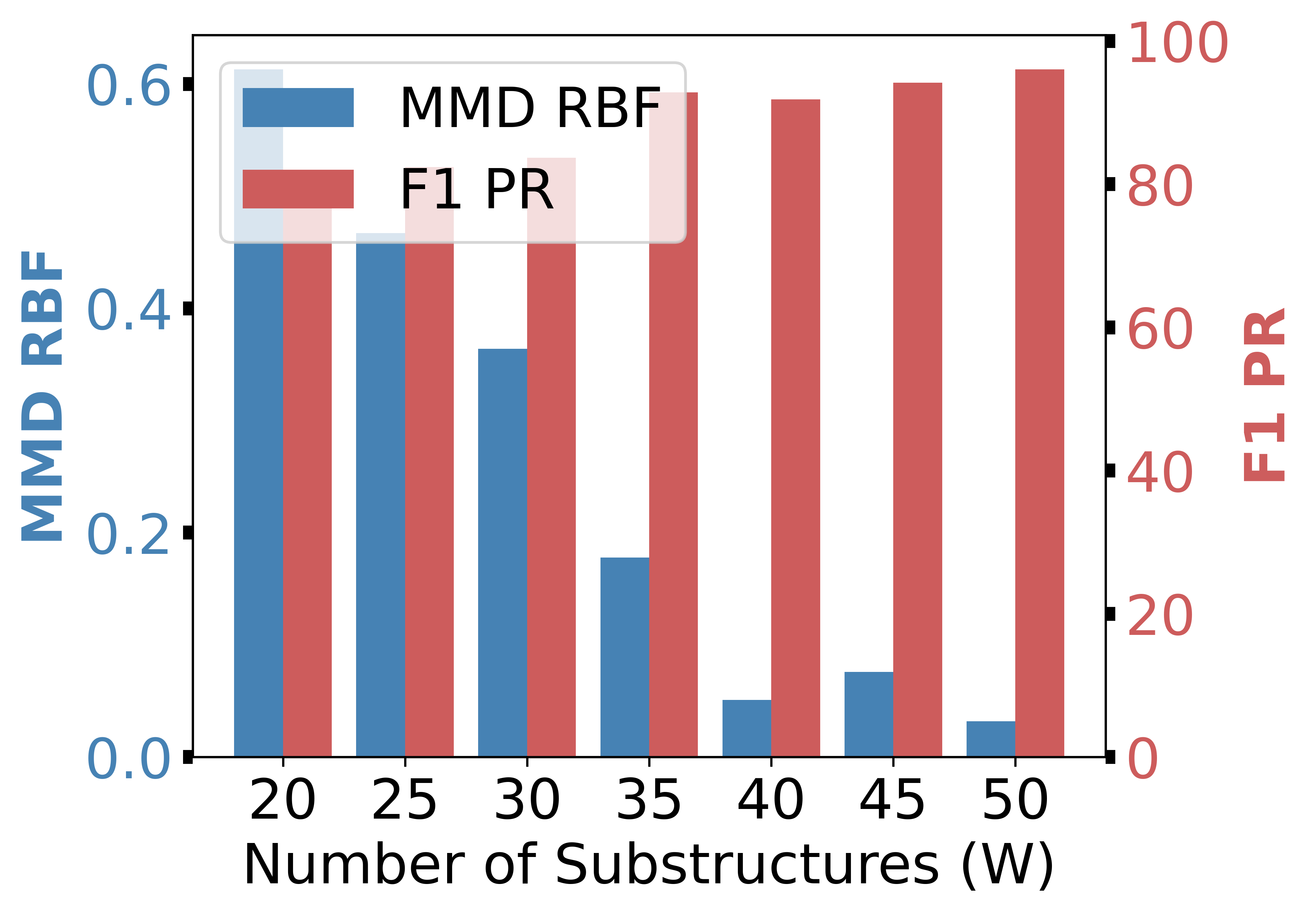}
  }
  \subfigure[Topic Count ($K$)]{
      \includegraphics[width=0.3\textwidth]{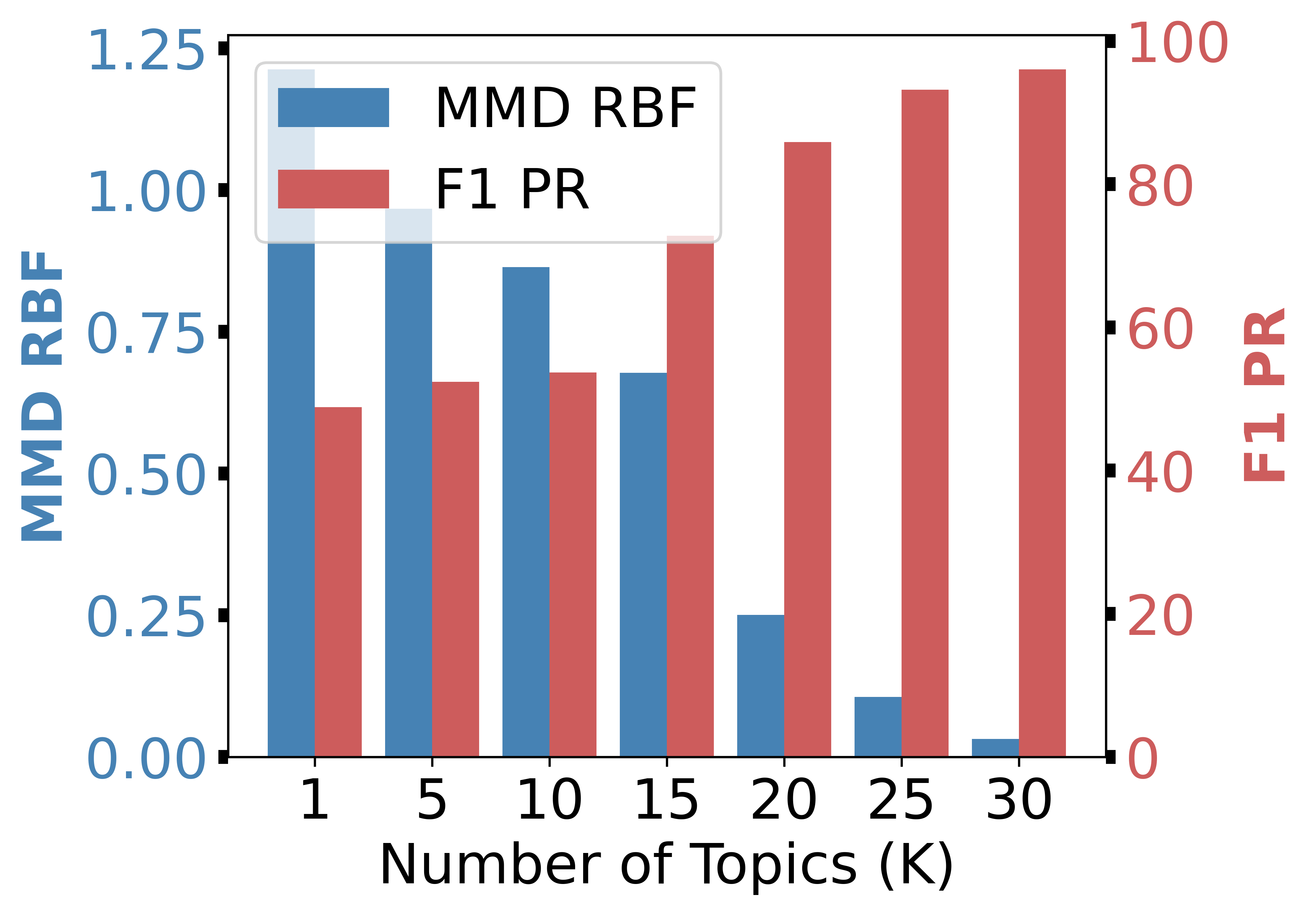}
  }
  \subfigure[Substructure Node Size ($n$)]{
      \includegraphics[width=0.3\textwidth]{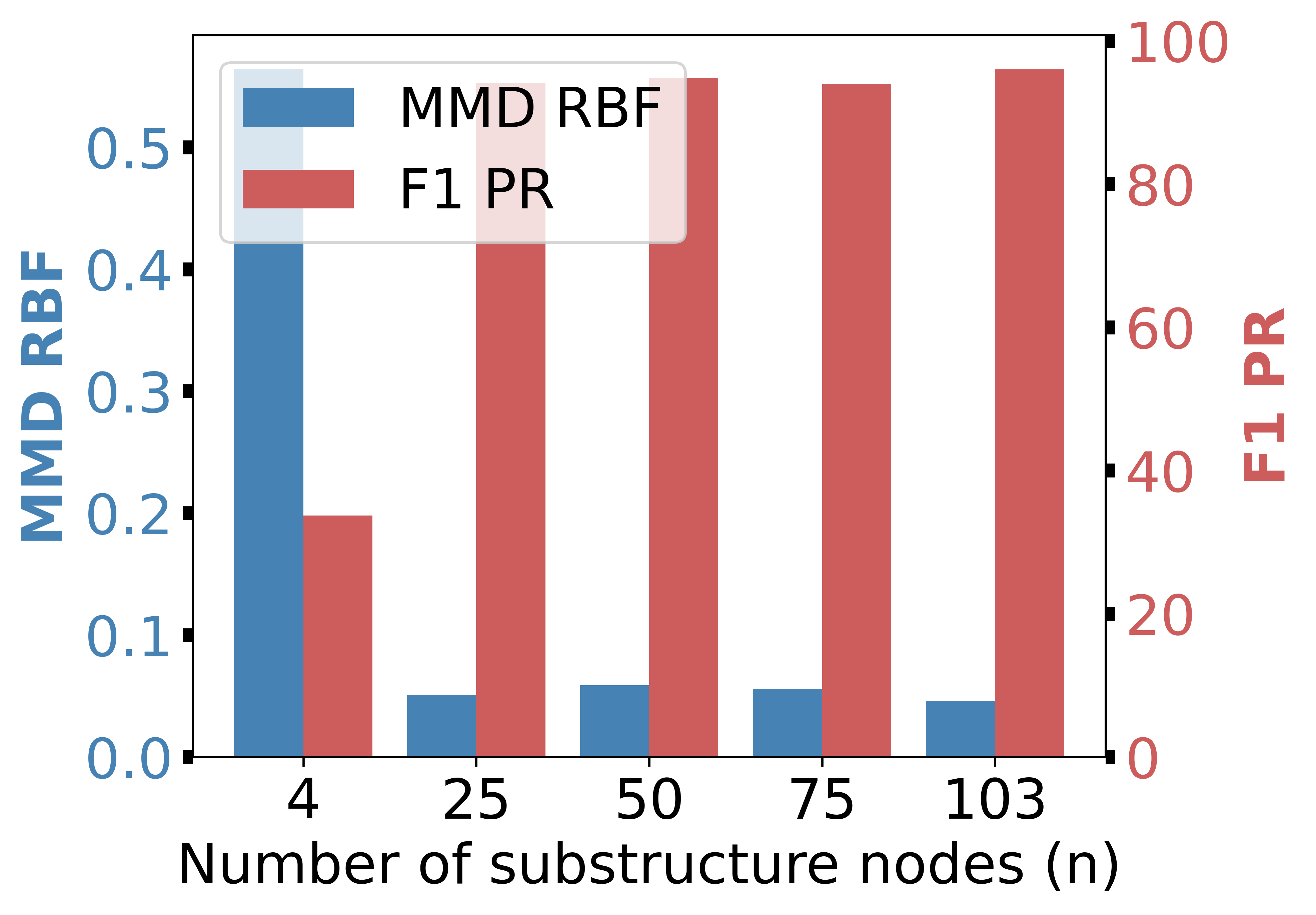}
  }
  \vspace{-7pt}
  \caption{Impact of key parameters on the PTC dataset. (a) Effect of substructure count ($W$). (b) Effect of topic count ($K$). (c) Effect of substructure node size ($n$).}
  \label{fig:PTC_params}
  \vspace{-10pt}
\end{figure}

\subsubsection{Parameter Sensitivity}
Figure~\ref{fig:PTC_params} illustrates how varying key parameters affects graph generation performance on the PTC dataset. The results reveal consistent trends across both diversity (F1 PR) and realism (MMD RBF) metrics.
(a) Substructure Count $W$: Increasing the number of sampled substructures significantly enhances both graph realism and diversity. MMD RBF decreases sharply while F1 PR approaches saturation as $W$ increases from 20 to 50. This suggests that denser substructure sampling enables NGTM to better capture nuanced structural variations.
(b) Topic Count $K$: Expanding the number of latent topics from 1 to 30 results in a marked improvement in performance. As $K$ grows, the model gains greater flexibility in organizing substructures into semantically coherent patterns. 
(c) Substructure Node Size $n$: Larger substructure sizes improve performance, especially in realism. When $n$ increases from 4 to 25, MMD RBF drops by over 80\%, indicating that small substructures may fail to capture essential motifs. 
These trends collectively indicate that a careful balance of these parameters is essential not only for optimizing the expressiveness and quality of the generated graphs, but also for maintaining a trade-off between performance and interpretability.

\begin{figure}
  \centering
  \includegraphics[width=0.9\textwidth]{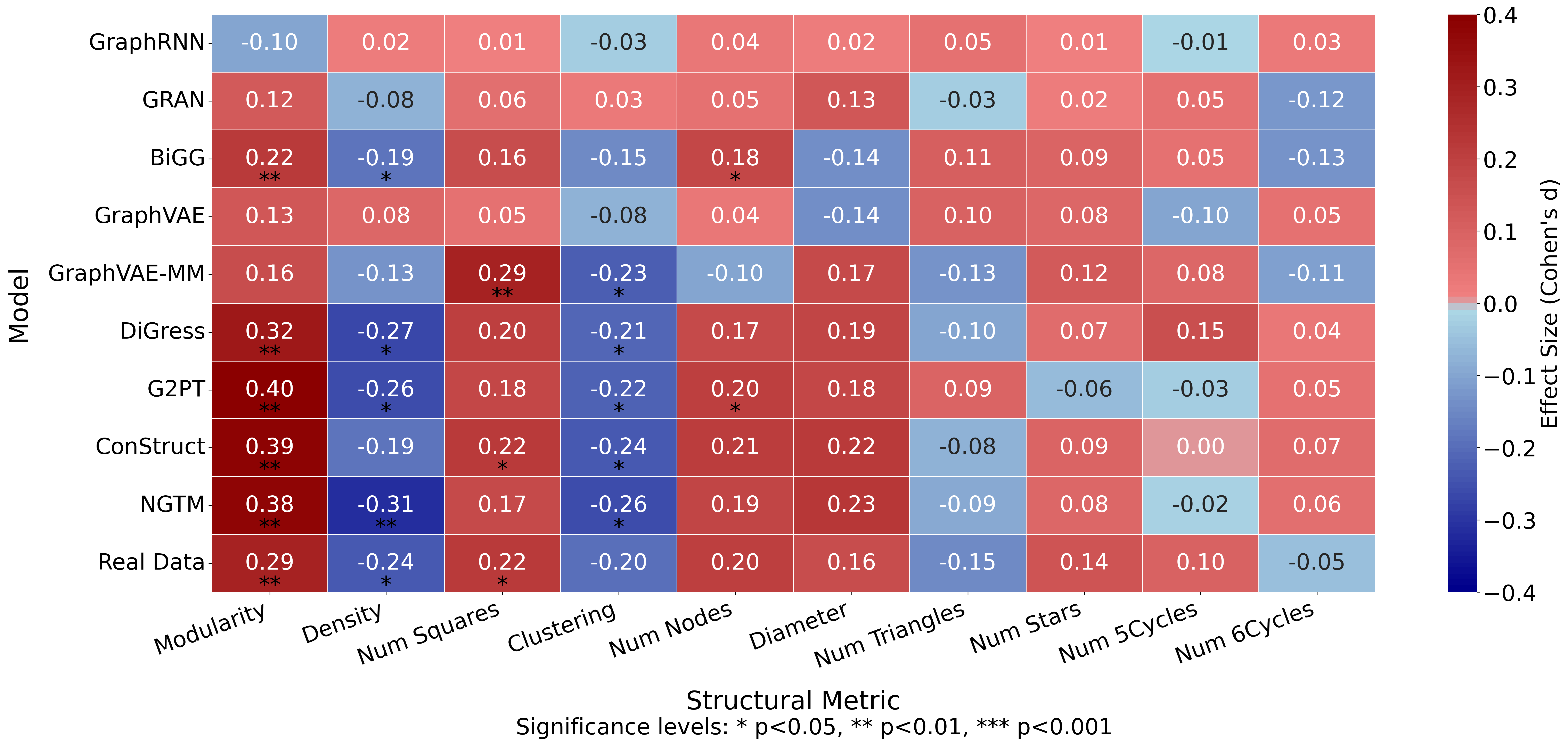}
  \vspace{-10pt}
  \caption{Comparison of Structural Metric Effect Sizes Between Real and Generated Graphs across Models.}
  \label{fig:ptc_effects}
  \vspace{-10pt}
\end{figure}

\subsubsection{Evaluation on Capturing Class-wise Structural Differences}
To evaluate whether NGTM can effectively capture semantically meaningful structural differences between graph classes, we design an experiment focused on distinguishing carcinogenic from non-carcinogenic graphs using the PTC dataset. PTC provides real-world binary labels for molecular graphs, where structural properties are known to strongly correlate with biological functions~\cite{gonzalez2001application, swamidass2005kernels, blinova2003toxicology}. We examine ten structural metrics—including modularity, density, numbers of cycles (4-cycles, 5-cycles, 6-cycles), clustering coefficient, diameter, and node count—which collectively capture key aspects of molecular topology such as sparsity, cyclicity, and community structure, all critical for understanding carcinogenic behavior.

We train a GIN classifier on real PTC graphs and apply it to graphs generated by NGTM. To assess how well NGTM preserves class-specific structural differences, we compute Cohen’s d effect sizes and corresponding p-values for each metric. Cohen’s d quantifies the difference between the average values of the two classes, normalized by their internal variability: larger absolute values indicate stronger separation between classes, while the sign indicates which class has higher scores. P-values, obtained from standard two-sample t-tests, measure how likely such a difference would arise by chance if there were no true difference; smaller values (e.g., p<0.05) suggest statistical significance.

Figure~\ref{fig:ptc_effects} summarizes the results across all models. Real PTC data (bottom row) shows strong structural differentiation between classes, including higher modularity and more 4-cycles, but lower density and clustering in carcinogenic graphs—consistent with known biological patterns~\cite{gonzalez2001application, swamidass2005kernels, blinova2003toxicology}. Among all generative models, NGTM best replicates the effect size patterns observed in real data, closely matching both the direction and magnitude of class-specific structural trends across most metrics. Notably, NGTM is the only model that consistently captures the full profile of structural effects, including significant separation on modularity, density, and 4-cycles.
These results demonstrate that NGTM not only produces realistic graphs but also captures meaningful structural variations associated with biological classes, supporting the model’s interpretability and semantic alignment with domain knowledge.

\subsection{Analysis of Learned Substructure Topics}

To further verify that NGTM captures meaningful and interpretable structural patterns, we conduct an analysis of the learned substructure topics. We select the PTC dataset for this analysis because it offers real-world binary labels and well-documented structural characteristics, making it a strong testbed for evaluating whether learned topics align with known biological properties~\cite{blinova2003toxicology}. Notably, we use a model trained with five latent topics to ease the presentation of the effects of different topics in a model.
Specifically, we organize the analysis into three parts:
First, we sample and visualize representative substructures from each latent topic to understand their semantic meaning.
Second, we study how varying topic weights impact key graph-level structural properties.
Finally, we examine how topic manipulations affect the predicted biological class (carcinogenic vs. non-carcinogenic), validating the biological relevance of the learned topics.

\begin{figure}[]
  \centering
  \includegraphics[width=0.85\textwidth]{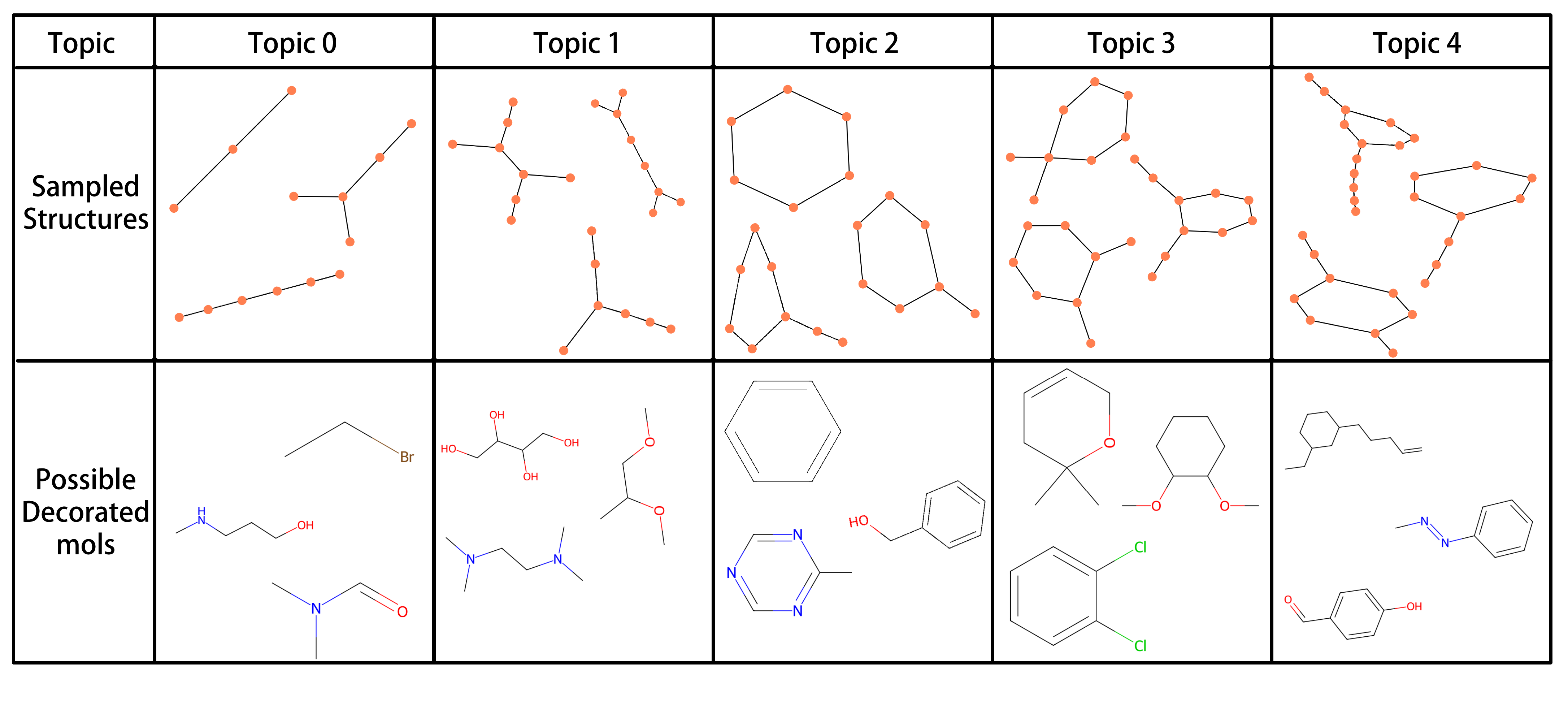}
  \vspace{-10pt}
  \caption{Sampled substructure skeletons (top row) from each NGTM topic, and corresponding chemically plausible molecules (bottom row) generated by decorating the skeletons based on PTC dataset hypotheses. }
  \label{fig:topic_structures}
  \vspace{-10pt}
\end{figure}

\subsubsection{Topic Visualization}
We conduct a topic visualization analysis for a better understanding of the semantic meanings captured by the learned topics. We first randomly sample three latent vectors from each topic distribution and decode them using the NGTM substructure decoder to obtain representative substructure skeletons (top row of Figure~\ref{fig:topic_structures}). These skeletons serve as canonical examples of the structural motifs captured by each topic.
To interpret these structures, we propose chemically plausible decorated molecules. Specifically, we reference the Positive and Negative Hypotheses extracted from the PTC dataset~\cite{blinova2003toxicology}, which describe substructures associated with carcinogenic (Positive) and non-carcinogenic (Negative) classes. By matching sampled skeletons to these hypotheses, we generate decorated molecules (bottom row of Figure~\ref{fig:topic_structures}) that ground the abstract motifs in real-world chemical knowledge.

The sampled structures reveal clear and coherent semantic distinctions between topics, consistent with the class-wise structural trends observed earlier (Figure~\ref{fig:ptc_effects}). Topic 0 predominantly generates linear or lightly branched chains, resembling simple alkyl structures typical of non-carcinogenic compounds. Topic 1 produces star-like tree structures with short diameters, similarly favoring benign graphs. Topic 2 is dominated by small ring systems and fused cycles, representing a structurally neutral scaffold family. Topic 3 generates extended polycyclic frameworks that are associated with high modularity and increased carcinogenic probability. Topic 4 produces irregular and ladder-like cyclic structures, showing a mild alignment toward carcinogenicity.
This visualization demonstrates that NGTM’s latent topics not only partition the graph space meaningfully but also align well with chemical motifs relevant to biological function.

\begin{figure}[]
  \centering
  \includegraphics[width=0.85\textwidth]{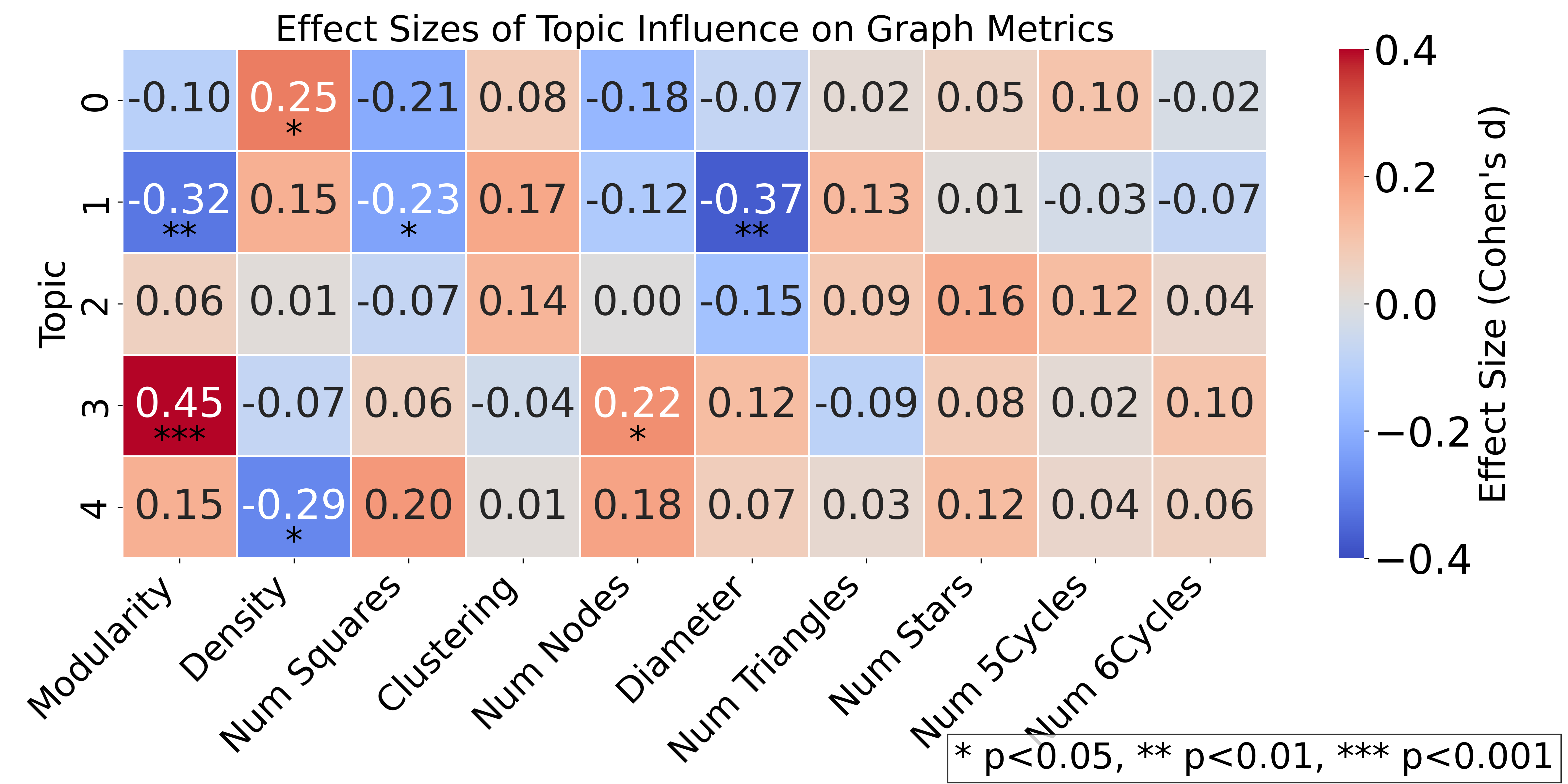}
  \vspace{-10pt}
  \caption{Topic-wise impacts on structural metrics evaluated through systematic topic weight manipulation.}
  \label{fig:topic_effects_a}
  \vspace{-10pt}
\end{figure}

\subsubsection{Influence of Topic Manipulation on Graph Structure}
We conduct a controlled topic manipulation experiment to evaluate whether the learned topics meaningfully control graph-level properties. This complements earlier findings on class-specific structural differences (Figure~\ref{fig:ptc_effects}) by directly testing the causal relationship between topic proportions and generated graph structures.
In this experiment, we progressively increase the weight of a target topic by adjusting the topic mixture vector $\theta$: specifically, for each topic $T_i$, we shift its normalized weight from –0.30 to +0.50 in increments of 0.15, while proportionally rescaling the weights of the remaining topics to maintain a valid probability distribution. For each setting, we generate 300 graphs to compute structural metrics. Changes are quantified using Cohen’s d effect sizes, and statistical significance is assessed via standard two-sample t-tests.

Figure~\ref{fig:topic_effects_a} shows the effect sizes of each topic across ten structural metrics. Warmer colors indicate positive shifts, cooler tones represent decreases, and asterisks denote statistical significance.
Among all topics, Topic 3 exhibits the strongest influence: increasing its weight substantially raises modularity, accompanied by moderate increases in node count and diameter. This matches the known profile of carcinogenic graphs, which tend to be large, modular, and sparse~\cite{swamidass2005kernels, gonzalez2001application}, suggesting that Topic 3 encodes carcinogenic-like structural motifs. In contrast, Topic 1 shows an opposite trend: it reduces modularity and diameter, promoting compact, highly connected graphs—a characteristic associated with non-carcinogenic structures~\cite{blinova2003toxicology}.
Other topics exhibit more localized effects. Topic 0 increases graph density and the number of 4-cycles, while slightly reducing clustering, consistent with heteroaromatic scaffolds containing square-like motifs. Topic 4 decreases density but increases clustering, triangles, and 6-cycles, resembling the structural patterns of polycyclic natural products. Topic 2, in contrast, shows minimal and statistically insignificant effects across most metrics, indicating it may serve as a neutral or background topic.

\begin{figure}[]
  \centering
  \subfigure[Reducing Graph Diameter via Topic 1 Adjustment]{
    \includegraphics[width=0.48\textwidth]{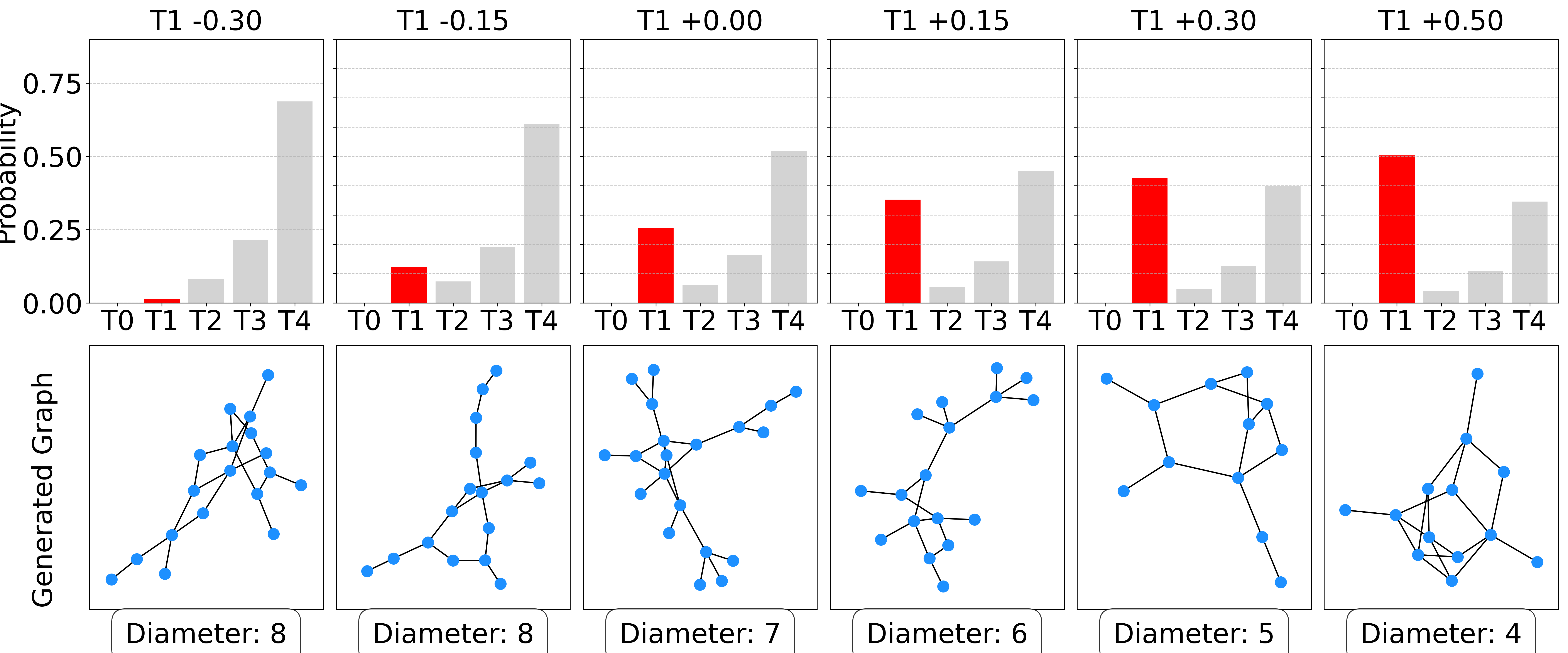}
  }
  \subfigure[Increasing Graph Modularity via Topic 3 Adjustment]{
    \includegraphics[width=0.48\textwidth]{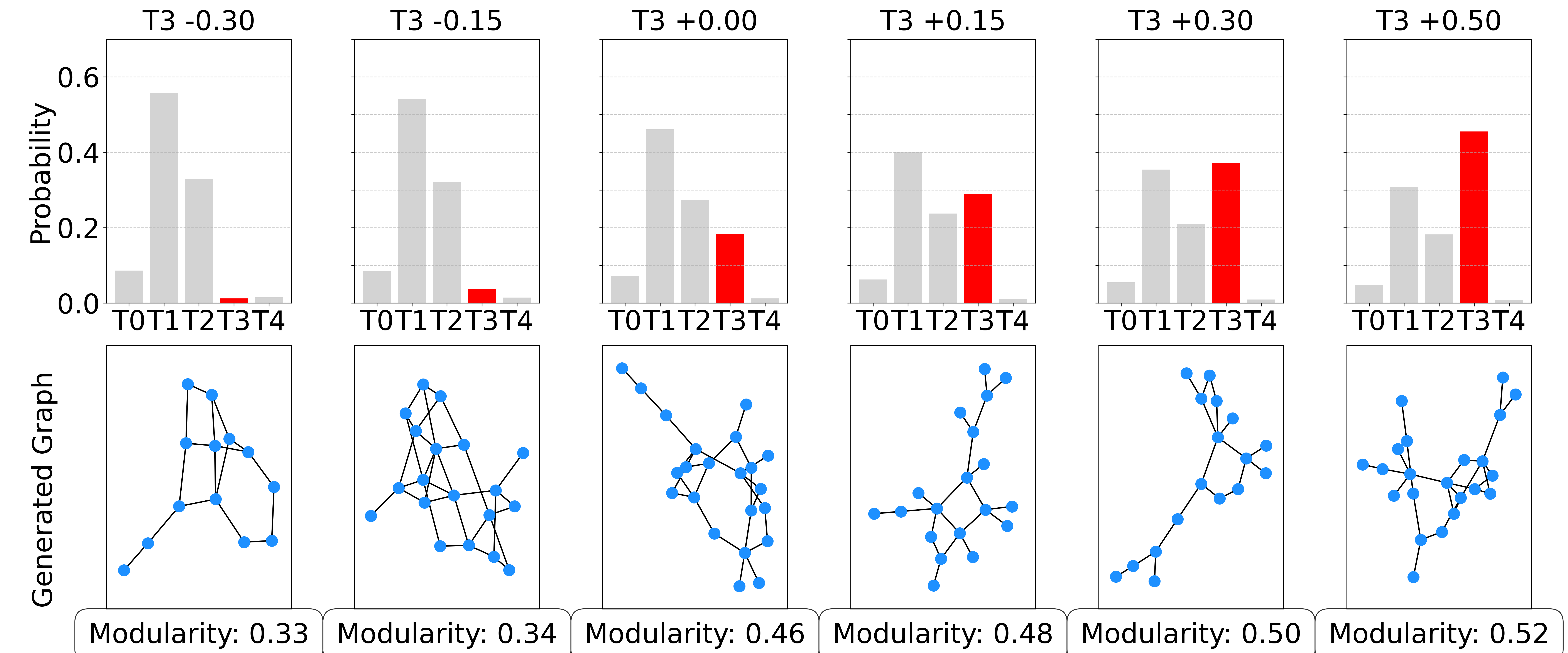}
  }
  \vspace{-10pt}
  \caption{Structural changes induced by progressively increasing the contribution of Topic 1 and Topic 3.}
  \label{fig:topic_effects_bc}
  \vspace{-10pt}
\end{figure}

Figures~\ref{fig:topic_effects_bc}(a) and \ref{fig:topic_effects_bc}(b) illustrate concrete examples of topic manipulation in NGTM. In Figure~\ref{fig:topic_effects_bc}(a), progressively increasing the weight of Topic 1 shrinks the diameter of generated graphs from 8 to 4 nodes, further reinforcing its strong link to compact topologies. Conversely, in Figure~\ref{fig:topic_effects_bc}(b), increasing the weight of Topic 3 steadily boosts modularity from 0.33 to 0.52, reflecting the emergence of more pronounced multi-community structures.
These findings confirm that NGTM’s learned topics act as interpretable and controllable factors that systematically influence global graph properties. Moreover, the observed trends are consistent with class-specific structural patterns identified in real PTC graphs, reinforcing the semantic alignment and interpretability of NGTM. Additional examples of topic manipulation can be found in Appendix Figure~\ref{fig:collection}.

\begin{figure}[]
  \centering
  \includegraphics[width=0.95\textwidth]{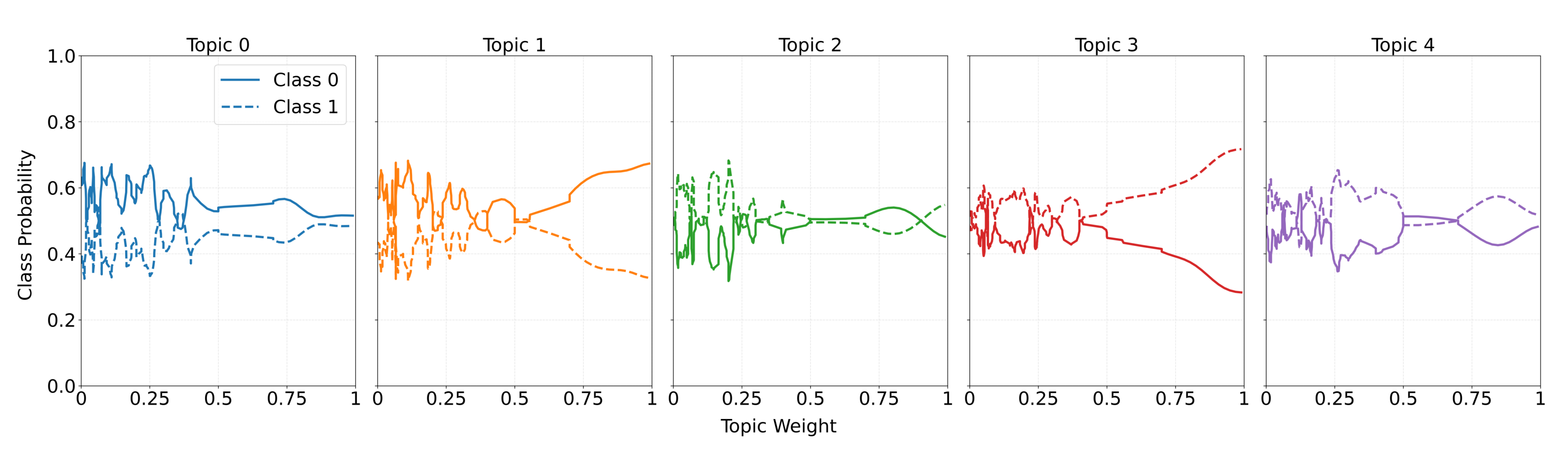}
  \vspace{-10pt}
  \caption{Effect of progressively increasing individual topic weights on the class probability of generated graphs in the PTC dataset. Solid lines represent Class 0 (non-carcinogenic) and dashed lines represent Class 1 (carcinogenic).}
  \label{fig:topic_class_prob}
  \vspace{-10pt}
\end{figure}

\subsubsection{Influence of Topic Manipulation on Class Probabilities}
To further validate whether NGTM’s learned topics align with meaningful biological properties, we design an experiment to examine how manipulating the weight of individual topics affects the predicted graph class (carcinogenic vs. non-carcinogenic). This complements earlier structural analyses (Figure~\ref{fig:ptc_effects}) by testing whether the structural shifts induced by topic manipulation translate into corresponding shifts in biological classification, thus providing behavioral evidence for the semantic grounding of the learned topics.
Specifically, for each latent topic, we progressively vary its normalized weight from 0 to 1 in increments of 0.05, while proportionally decreasing the weights of the remaining topics to maintain a valid mixture (i.e., the topic proportions always sum to 1). At each setting, we generate 300 graphs and predict their classes using a GIN classifier trained on real PTC graphs. We then compute the proportion of generated graphs classified as carcinogenic (Class 1) or non-carcinogenic (Class 0) under each topic manipulation.

Figure~\ref{fig:topic_class_prob} shows the results. Solid lines represent the probability of being classified as Class 0 (non-carcinogenic), while dashed lines represent Class 1 (carcinogenic).
The findings are consistent with the structural patterns observed earlier: Topic 3 strongly promotes carcinogenicity: as its weight increases, the Class 1 probability rises monotonically, reaching about 0.75 when Topic 3 dominates the mixture. This is coherent with Figure~\ref{fig:ptc_effects}, where Topic 3 was shown to boost modularity and graph size—two key traits associated with carcinogenic graphs~\cite{swamidass2005kernels, gonzalez2001application}. Topic 1 exhibits the opposite effect: increasing its weight enhances the likelihood of Class 0, indicating a shift toward benign, compact structures. This matches its role in producing graphs with reduced diameter and modularity. Topic 0 slightly favors Class 0, likely due to its influence on increasing density and 4-cycle motifs, structural features often found in non-carcinogenic compounds. Topic 4 mildly favors Class 1, consistent with its enrichment of triangle motifs and 6-cycles—patterns associated with certain polycyclic frameworks found in carcinogenic molecules. Topic 2 remains largely neutral, corroborating its weak and statistically insignificant influence across structural metrics seen previously.
These results reinforce that NGTM’s topics do not merely partition graph structures randomly but capture interpretable, biologically relevant factors. The directional shifts in class probability are fully consistent with prior Cohen’s d analyses of structural metrics, confirming that the semantic meaning embedded in each topic manifests both in structure and biological function.

\begin{figure}[htbp]
  \centering
  \includegraphics[width=1\textwidth]{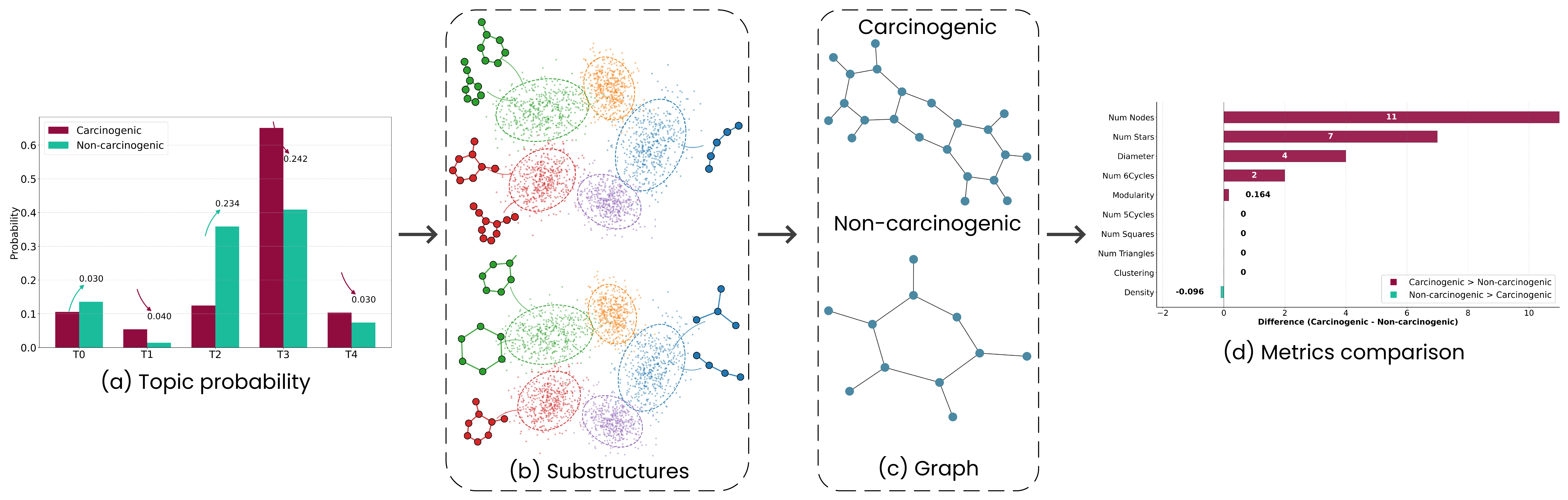}
  \vspace{-8pt}
  \caption{Interpretable Comparison of Carcinogenic and Non-carcinogenic Graphs Generated by NGTM.}
  \label{fig:case}
  \vspace{-8pt}
\end{figure}

\subsection{Case Study: Interpretable Comparison of Carcinogenic and Non-carcinogenic Graphs}

Figure~\ref{fig:case} presents an interpretable comparison between carcinogenic and non-carcinogenic graphs, illustrating how NGTM’s topic-driven generation process reflects biologically meaningful structural differences. In (a), we show the topic proportion vectors sampled for each class. The carcinogenic graph exhibits a dominant weight on Topic 3, whereas the non-carcinogenic graph places greater emphasis on Topic 2. These proportions determine the selection of substructures from the learned topic distributions. As shown in (b), the carcinogenic graph incorporates more polycyclic and modular motifs (red and green regions), while the non-carcinogenic graph emphasizes compact and simpler motifs such as rings and chains. These sampled substructures are then assembled into full graphs, shown in (c), where the carcinogenic structure is visibly larger and more modular, while the non-carcinogenic structure is smaller and denser. The resulting structural properties, summarized in (d), confirm this distinction: the carcinogenic graph has more nodes, higher diameter, and more star motifs, while the non-carcinogenic graph shows slightly higher density. These observations are consistent with previous findings (e.g., Figures~\ref{fig:ptc_effects} and~\ref{fig:topic_effects_a}), where Topic 3 was linked to traits such as increased modularity and carcinogenicity, and Topic 2 was associated with compact, neutral structures. This case study demonstrates that NGTM not only generates graphs with realistic topologies but also aligns its topic-driven compositional process with interpretable, class-specific semantics, reinforcing its potential for biologically informed graph generation.

\section{Discussion}

NGTM introduces a new paradigm for interpretable graph generation by modeling graphs as mixtures of latent substructure topics. Evaluated across four benchmark datasets, NGTM not only achieves competitive generation quality but also provides a transparent and controllable generative process. By linking distinct structural patterns such as modularity, diameter and cyclicity to separate latent topics, NGTM allows users to fine-tune both global graph properties and local substructures through intuitive topic manipulation.
This fine-grained control is promising for applications like molecular design, where adjusting structural features can steer the generation of molecules toward desired properties, such as lower toxicity or improved bioavailability. Indeed, NGTM’s interpretable topic-based framework opens avenues for broader scientific and engineering tasks. In materials science, for instance, guiding the formation of specific crystalline motifs could accelerate the discovery of novel materials. In knowledge graph construction, controlling the emergence of particular relational motifs may enhance logical coherence and downstream reasoning capabilities.

Future extensions of NGTM could explore hierarchical topic structures, where topics themselves are organized at multiple levels of abstraction, enabling even richer semantic modeling. Integrating property-aware training objectives would further enable goal-directed graph generation, where users can specify desired graph properties alongside topic distributions. Additionally, coupling NGTM with expert-in-the-loop interfaces—allowing users to edit, refine, or create new topics based on domain expertise—could make the model even more powerful for scientific discovery and design applications.


\section{Methods}

In this section, we begin with backgrounds on graph generative models and topic modeling. We then describe the model architecture, and finally detail the inference procedure and the loss function.

\subsection{Backgrounds}

\textbf{Graph Generation }
Over the past decade, significant efforts have been dedicated to advancing graph generation techniques, including generative adversarial networks (GANs)~\cite{de2018molgan, wang2023fairness}, variational auto-encoders (VAEs)~\cite{GraphVAE, GraphVAE–MM, hou2024dag}, autoregressive models (ARs)~\cite{kong2023autoregressive, bu2023let}, normalizing flows (NFs)~\cite{luo2021graphdf, kuznetsov2021molgrow}, and diffusion models (DMs)~\cite{vignac2022digress, cho2024multi, minello2024graph}. These models have provided robust frameworks for generating realistic and diverse graphs, with applications in molecular chemistry~\cite{luo2021graphdf, zang2020moflow} and program synthesis~\cite{brockschmidt2018generative, dai2018syntax}. However, many of these methods function as black boxes, making it difficult to understand or interpret how specific graph structures are generated.
Existing efforts toward interpretability in graph generation have primarily focused on disentanglement-based methods~\cite{stoehr2019disentangling, guo2020interpretable, li2020dirichlet, du2022interpretable}, which aim to learn latent factors aligned with controllable aspects of graphs, such as node or edge attributes. These approaches provide insight into global properties but often fail to explain the step-by-step structure-building process. In parallel, some methods explore substructure-based generation. JT-VAE~\cite{jin2018junction} assembles molecules from chemically valid fragments using junction trees, while PS-VAE~\cite{kong2022principal} leverages frequently occurring subgraphs as building blocks. Although these methods enhance generation quality and modularity, they are not designed for interpretability and do not explain why or how substructures are selected and composed.
In contrast, NGTM uniquely introduces a topic modeling perspective that organizes substructures into semantically meaningful topics. This provides not only local interpretability through explicit substructures but also global coherence through interpretable topic distributions. Our model offers a transparent generative narrative: each graph is generated by first sampling topic mixtures (explaining the  ``why''), then selecting substructures governed by these topics (explaining the  ``what''), and finally assembling them under structural guidance (explaining the  ``how'').

\textbf{Topic Modeling and Neural Topic Models }
Topic modeling is a foundational technique in natural language processing (NLP), originally developed to uncover latent semantic structures within large text corpora. Classical topic models, such as Latent Dirichlet Allocation (LDA)~\cite{blei2003latent}, Probabilistic Latent Semantic Analysis (PLSA)~\cite{hofmann2001unsupervised}, and Non-negative Matrix Factorization (NMF)~\cite{lee2000algorithms}, treat documents as mixtures of topics, where each topic corresponds to a distribution over words. These models provide interpretable insights by identifying recurring patterns of word co-occurrence that align with intuitive semantic themes.
To overcome the limited expressivity and scalability of classical models, neural topic models (NTMs) have been proposed. Representative works include the Neural Variational Document Model (NVDM)~\cite{miao2016neural}, Neural Variational Inference for Topic Models (NVITM)~\cite{miao2017discovering}, and contextualized topic models using deep encoders~\cite{bianchi2020pre}. These methods integrate variational inference with neural networks to learn flexible, high-capacity topic representations.

Although widely adopted in textual domains, topic modeling has been underexplored in the context of graph generation. Most prior work involving topic models for graphs focuses on community detection or node classification, treating the graph as static input rather than a generative target~\cite{long2020graph, xie2021graph}. In contrast, we argue that graphs—particularly molecular graphs—exhibit structure that is naturally compositional and semantically organized, making them well-suited for a topic modeling framework. Substructures in graphs (e.g., rings, chains, motifs) can be seen as the graph analog of ``words'' and entire graphs resemble ``documents'' composed from mixtures of such semantic components.

Our proposed NGTM framework explicitly adopts this analogy by modeling graphs as mixtures of latent substructure topics. Each topic captures a distribution over interpretable substructures, and the generative process assembles graphs by sampling from these topics under global guidance. This structured approach enables both semantic transparency and traceability throughout generation. To the best of our knowledge, NGTM is the first framework that integrates topic modeling principles into graph generation to achieve both structural quality and interpretability.

\begin{figure*}
  \centering
  \includegraphics[width=1\textwidth]{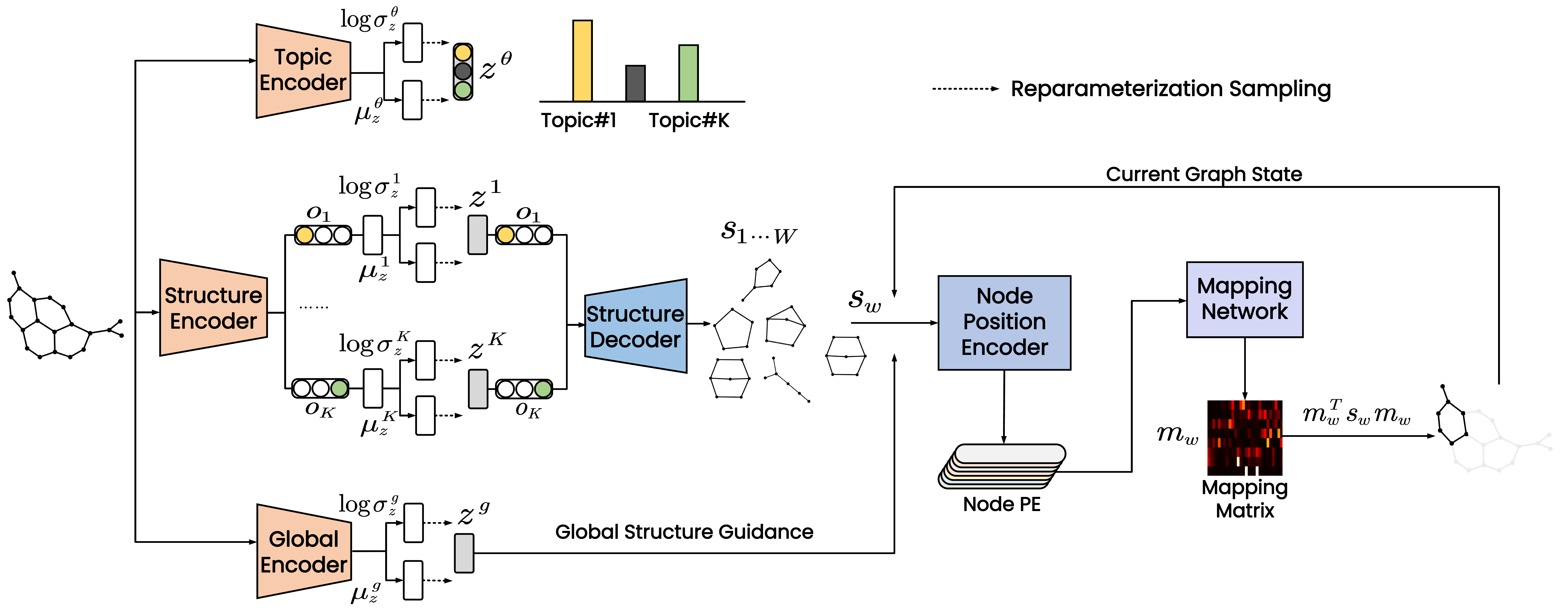}
  \vspace{-15pt}
  \caption{The Overview of NGTM Architecture.}
  \label{fig:model}
  \vspace{-10pt}
\end{figure*}

\subsection{Modular Architecture of NGTM}

The NGTM framework is designed to learn topic distributions, substructure distributions, and a global structural prior from training data, while also learning how to assemble substructures into complete graphs.
As illustrated in Figure~\ref{fig:model}, NGTM consists of several key components: the Topic Encoder, Structure Encoder, Global Encoder, Structure Decoder, Node Position Encoder, and Mapping Network.

\textbf{Encoder Module:}
NGTM uses three encoders to extract topic semantics, substructure-level patterns, and global structure context. The Topic Encoder and Global Encoder process the full graph G and output the parameters of corresponding Gaussian distributions via MLPs:
\[\mu_z^\theta, \sigma_z^\theta = \text{MLP}(\text{Enc}_\text{Topic}(G)), \quad \mu_z^g, \sigma_z^g = \text{MLP}(\text{Enc}_\text{Global}(G)).\]
To model topic-specific substructure priors, NGTM employs Conditional VAEs~\cite{sonderby2016ladder}. Each graph G is encoded by the Structure Encoder and concatenated with a one-hot topic vector $o_k$. The result is passed through MLPs to produce:
\[\mu_{z}^k, \sigma_{z}^k = \text{MLP}(\text{Enc}_\text{Structure}(G) \oplus o_k).\]
Here, the topic vector acts as a control signal, guiding the model to learn distinct distributions over structural motifs for each topic. The reparameterization trick~\cite{kingma2013auto} enables gradient-based optimization.

\textbf{Decoder Module:}
Each sampled substructure vector $z^{k = c_w}$ is concatenated with its topic one-hot vector and decoded as follows:
\[s_w = \text{StructureDecoder}(z^{k = c_w} \oplus o_{k = c_w}).\]
This conditioning ensures that each generated substructure is semantically linked to a specific topic, enhancing interpretability.

\textbf{Substructure Assembly Module:}
Each substructure $s_w$ consists of up to $n$ nodes and is represented by an adjacency matrix $A_{s_w}$. The graph is constructed by iteratively assembling $W$ substructures. At each step $w$, the current graph state $G_{w-1}$, the substructure $s_w$, and the global structure vector $g$ are input into the Node Position Encoder to produce a Node Position Encoding:
\[p_w = \text{Node Position Encoder}(s_w, G_{w-1}, g).\]
This encoding is fed into the Mapping Network to generate a softmax-normalized mapping matrix:
\[m_w = \text{Softmax}(\text{Mapping Network}(p_w)).\]
The matrix $m_w$ softly aligns nodes in $s_w$ with positions in the current graph. The adjacency matrix is updated as:
\[A_w = A_{w-1} + m_w^{T} A_{s_w} m_w.\]
This soft assignment enables flexible integration of substructures while preserving a traceable correspondence between each motif and its position in the graph, reinforcing transparency throughout the generation process.

\subsection{Inference and Optimization}

To enable efficient and interpretable learning, NGTM introduces a set of latent variables $Z = \{z^{\theta}, z^{1}, \ldots, z^{K}, z^g\}$ , capturing topic proportions, substructure semantics, and global graph characteristics. These variables are modeled as samples from Gaussian distributions:
\[
z^\theta \sim \mathcal{N}(\mu_z^\theta, \sigma_z^\theta), \quad z^g \sim \mathcal{N}(\mu_z^g, \sigma_z^g), \quad z^k \sim \mathcal{N}(\mu_z^k, \sigma_z^k), \; \text{for } k = 1, \ldots, K.
\]
We optimize the model parameters \(\phi\) using variational inference~\cite{blei2017variational}, which maximizes the Evidence Lower Bound (ELBO) on the log-likelihood of a graph $G$:
\begin{align}
\log p(G) \geq \mathrm{ELBO} &= \mathbb{E}_{q_\phi(Z \mid G)} \left[ \log p(G \mid Z) \right] \nonumber - D_{\mathrm{KL}}\left[q_\phi(Z \mid G) \,\|\, p(Z)\right].
\label{eq:2}
\end{align}
The first term encourages faithful reconstruction of the input graph, while the second term regularizes the posterior distribution by penalizing deviation from the prior.
We assume conditional independence among latent variables, enabling factorization of both the variational posterior and prior:
\[
q_\phi(Z \mid G) = q_\phi(z^\theta \mid G) \prod_{k=1}^{K} q_\phi(z^k \mid G) \cdot q_\phi(z^g \mid G),
\]
\[
p(Z) = p(z^\theta) \prod_{k=1}^{K} p(z^k) \cdot p(z^g).
\]
Under this factorization, the KL divergence decomposes as:
\begin{align}
D_{\mathrm{KL}}[q_\phi(Z \mid G) \| p(Z)] &= D_{\mathrm{KL}}[q_\phi(z^{\theta} \mid G) \| p(z^{\theta})] \nonumber \\
&+ \sum_{k=1}^{K} D_{\mathrm{KL}}[q_\phi(z^{k} \mid G) \| p(z^{k})] \nonumber + D_{\mathrm{KL}}[q_\phi(z^{g} \mid G) \| p(z^{g})].
\label{eq:3}
\end{align}

\subsubsection{Loss Function}

The overall training objective of NGTM consists of a reconstruction loss, $(K + 2)$ KL divergence terms, and an orthogonality regularization term.

\textbf{KL Divergences.}  
We adopt standard normal priors for the topic proportion and global structure variables. The corresponding KL terms are:
\[
\mathcal{L}_{\text{KL}}^{\theta} = D_{\mathrm{KL}}\left[\mathcal{N}(\mu_{z}^{\theta}, \sigma_{z}^{\theta}) \parallel \mathcal{N}(0, I)\right], \quad
\mathcal{L}_{\text{KL}}^{g} = D_{\mathrm{KL}}\left[\mathcal{N}(\mu_{z}^{g}, \sigma_{z}^{g}) \parallel \mathcal{N}(0, I)\right].
\]
For the $K$ topic-specific substructure distributions, we use learnable priors $\mathcal{N}(\mu^k, \sigma^k)$ to allow flexible modeling of diverse structural motifs. Each topic's KL divergence is weighted by its topic proportion $z_k^\theta$:
\[
\mathcal{L}_{\text{KL}}^{K} = \sum_{k=1}^{K} z^\theta_k \cdot D_{\mathrm{KL}}\left[\mathcal{N}(\mu_z^k, \sigma_z^k) \parallel \mathcal{N}(\mu^k, \sigma^k)\right].
\]

\textbf{Reconstruction Loss.}  
We adopt the micro-macro loss~\cite{GraphVAE–MM}, which captures both local and global graph properties. The micro component enforces fine-grained structural fidelity (e.g., node degrees, pairwise connections), while the macro component aligns global statistics such as degree distributions, triangle counts, and multi-step transition probabilities. This design ensures permutation invariance and improves robustness to node ordering.

\textbf{Orthogonality Regularization.}  
To encourage semantic diversity among the learned topics, we introduce an orthogonality regularization term on the topic means:
\[
\mathcal{L}_{\text{ortho}} = \sum_{i=1}^{K}\sum_{\substack{j=1 \\ j \neq i}}^{K} \left((\mu_z^{k_i})^\top \mu_z^{k_j}\right)^2.
\]
This term penalizes alignment between different topic centers, promoting topic disentanglement.

\textbf{Total Loss.}  
The final objective combines all components with balancing weights:
\[
  \mathcal{L}_{\text{total}} = \alpha \cdot \mathcal{L}_{\text{rec}} + \beta \cdot \mathcal{L}_{\text{KL}}^{\theta} + \delta \cdot \mathcal{L}_{\text{KL}}^{g} + \gamma \cdot \mathcal{L}_{\text{KL}}^{K} + \omega \cdot \mathcal{L}_{\text{ortho}}.
\]
Here, $\alpha$, $\beta$, $\delta$, $\gamma$, and $\omega$ are scalar hyperparameters that balance the contributions of the reconstruction loss, KL divergence terms, and orthogonality regularization.

\section{Conclusion}

We introduced NGTM, a novel framework for interpretable graph generation that models graphs as mixtures of latent substructure topics. By explicitly organizing meaningful substructures under topic distributions and guiding their assembly with a global structural variable, NGTM ensures semantic transparency at both local and global levels throughout the generation process.
Extensive experiments across diverse datasets demonstrate that NGTM not only achieves competitive generation quality but also provides clear, traceable interpretations of how structural components shape the final graph. This interpretability empowers users to understand, control, and steer graph generation, making NGTM especially valuable for applications where transparency and precision are critical.
By bridging the gap between structural fidelity and semantic interpretability, NGTM points toward a promising direction for next-generation graph generative models—where generated structures are not only realistic but also meaningfully explainable and controllable.

\section{Data Availability}

All datasets used in this study are publicly available. The MUTAG and PTC datasets are accessed via the dgl.data.GINDataset interface from the Deep Graph Library (DGL): \url{https://www.dgl.ai/dgl_docs/generated/dgl.data.GINDataset.html}. The Ogbg-molbbbp dataset is accessed via the DglGraphPropPredDataset interface provided by the Open Graph Benchmark: \url{https://github.com/snap-stanford/ogb/blob/master/ogb/graphproppred/dataset_dgl.py}. The Lobster dataset is synthetically generated using the random\_lobster function from the NetworkX library. It was first used as a benchmark in the Graph Recurrent Attention Network (GRAN) paper: \url{https://github.com/lrjconan/GRAN}, and has since been widely adopted in subsequent graph generation research.

\bibliography{sn-bibliography}

\begin{thebibliography}{10}
\expandafter\ifx\csname url\endcsname\relax
  \def\url#1{\texttt{#1}}\fi
\expandafter\ifx\csname urlprefix\endcsname\relax\def\urlprefix{URL }\fi
\providecommand{\bibinfo}[2]{#2}
\providecommand{\eprint}[2][]{\url{#2}}

\bibitem{tian2025leveraging}
\bibinfo{author}{Tian, Y.} \emph{et~al.}
\newblock \bibinfo{title}{Leveraging domain motif assembler for multi-objective, multi-domain and explainable molecular design}  (\bibinfo{year}{2025}).

\bibitem{wu2025construction}
\bibinfo{author}{Wu, W.}, \bibinfo{author}{Wen, C.}, \bibinfo{author}{Yuan, Q.}, \bibinfo{author}{Chen, Q.} \& \bibinfo{author}{Cao, Y.}
\newblock \bibinfo{title}{Construction and application of knowledge graph for construction accidents based on deep learning}.
\newblock \emph{\bibinfo{journal}{Engineering, construction and architectural management}} \textbf{\bibinfo{volume}{32}}, \bibinfo{pages}{1097--1121} (\bibinfo{year}{2025}).

\bibitem{rigoni2025rgcvae}
\bibinfo{author}{Rigoni, D.}, \bibinfo{author}{Navarin, N.} \& \bibinfo{author}{Sperduti, A.}
\newblock \bibinfo{title}{Rgcvae: Relational graph conditioned variational autoencoder for molecule design}.
\newblock \emph{\bibinfo{journal}{Machine Learning}} \textbf{\bibinfo{volume}{114}}, \bibinfo{pages}{47} (\bibinfo{year}{2025}).

\bibitem{kumar2025explainable}
\bibinfo{author}{Kumar, A.}, \bibinfo{author}{Hora, H.}, \bibinfo{author}{Rohilla, A.}, \bibinfo{author}{Kumar, P.} \& \bibinfo{author}{Gautam, R.}
\newblock \bibinfo{title}{Explainable artificial intelligence (xai) for healthcare: Enhancing transparency and trust}.
\newblock In \emph{\bibinfo{booktitle}{International Conference on Cognitive Computing and Cyber Physical Systems}}, \bibinfo{pages}{295--308} (\bibinfo{organization}{Springer}, \bibinfo{year}{2025}).

\bibitem{zheng2024application}
\bibinfo{author}{Zheng, L.} \emph{et~al.}
\newblock \bibinfo{title}{Application scenario-oriented molecule generation platform developed for drug discovery}.
\newblock \emph{\bibinfo{journal}{Methods}} \textbf{\bibinfo{volume}{222}}, \bibinfo{pages}{112--121} (\bibinfo{year}{2024}).

\bibitem{doshi2017towards}
\bibinfo{author}{Doshi-Velez, F.} \& \bibinfo{author}{Kim, B.}
\newblock \bibinfo{title}{Towards a rigorous science of interpretable machine learning}.
\newblock \emph{\bibinfo{journal}{arXiv preprint arXiv:1702.08608}}  (\bibinfo{year}{2017}).

\bibitem{holzinger2019causability}
\bibinfo{author}{Holzinger, A.}, \bibinfo{author}{Langs, G.}, \bibinfo{author}{Denk, H.}, \bibinfo{author}{Zatloukal, K.} \& \bibinfo{author}{M{\"u}ller, H.}
\newblock \bibinfo{title}{Causability and explainability of artificial intelligence in medicine}.
\newblock \emph{\bibinfo{journal}{Wiley Interdisciplinary Reviews: Data Mining and Knowledge Discovery}} \textbf{\bibinfo{volume}{9}}, \bibinfo{pages}{e1312} (\bibinfo{year}{2019}).

\bibitem{tjoa2020survey}
\bibinfo{author}{Tjoa, E.} \& \bibinfo{author}{Guan, C.}
\newblock \bibinfo{title}{A survey on explainable artificial intelligence (xai): Toward medical xai}.
\newblock \emph{\bibinfo{journal}{IEEE transactions on neural networks and learning systems}} \textbf{\bibinfo{volume}{32}}, \bibinfo{pages}{4793--4813} (\bibinfo{year}{2020}).

\bibitem{guo2022systematic}
\bibinfo{author}{Guo, X.} \& \bibinfo{author}{Zhao, L.}
\newblock \bibinfo{title}{A systematic survey on deep generative models for graph generation}.
\newblock \emph{\bibinfo{journal}{IEEE Transactions on Pattern Analysis and Machine Intelligence}} \textbf{\bibinfo{volume}{45}}, \bibinfo{pages}{5370--5390} (\bibinfo{year}{2022}).

\bibitem{zhang2023survey}
\bibinfo{author}{Zhang, M.} \emph{et~al.}
\newblock \bibinfo{title}{A survey on graph diffusion models: Generative ai in science for molecule, protein and material}.
\newblock \emph{\bibinfo{journal}{arXiv preprint arXiv:2304.01565}}  (\bibinfo{year}{2023}).

\bibitem{liu2023generative}
\bibinfo{author}{Liu, C.} \emph{et~al.}
\newblock \bibinfo{title}{Generative diffusion models on graphs: Methods and applications}.
\newblock \emph{\bibinfo{journal}{arXiv preprint arXiv:2302.02591}}  (\bibinfo{year}{2023}).

\bibitem{chen2025graph}
\bibinfo{author}{Chen, X.} \emph{et~al.}
\newblock \bibinfo{title}{Graph generative pre-trained transformer}.
\newblock \emph{\bibinfo{journal}{arXiv preprint arXiv:2501.01073}}  (\bibinfo{year}{2025}).

\bibitem{wang2025exselfrl}
\bibinfo{author}{Wang, J.} \& \bibinfo{author}{Zhu, F.}
\newblock \bibinfo{title}{Exselfrl: An exploration-inspired self-supervised reinforcement learning approach to molecular generation}.
\newblock \emph{\bibinfo{journal}{Expert Systems with Applications}} \textbf{\bibinfo{volume}{260}}, \bibinfo{pages}{125410} (\bibinfo{year}{2025}).

\bibitem{hou2024dag}
\bibinfo{author}{Hou, D.}, \bibinfo{author}{Gao, C.}, \bibinfo{author}{Li, X.} \& \bibinfo{author}{Wang, Z.}
\newblock \bibinfo{title}{Dag-aware variational autoencoder for social propagation graph generation}.
\newblock In \emph{\bibinfo{booktitle}{Proceedings of the AAAI Conference on Artificial Intelligence}}, vol.~\bibinfo{volume}{38}, \bibinfo{pages}{8508--8516} (\bibinfo{year}{2024}).

\bibitem{xu2025generation}
\bibinfo{author}{Xu, C.}, \bibinfo{author}{Deng, X.}, \bibinfo{author}{Lu, Y.} \& \bibinfo{author}{Yu, P.}
\newblock \bibinfo{title}{Generation of molecular conformations using generative adversarial neural networks}.
\newblock \emph{\bibinfo{journal}{Digital Discovery}} \textbf{\bibinfo{volume}{4}}, \bibinfo{pages}{161--171} (\bibinfo{year}{2025}).

\bibitem{madeira2024generative}
\bibinfo{author}{Madeira, M.}, \bibinfo{author}{Vignac, C.}, \bibinfo{author}{Thanou, D.} \& \bibinfo{author}{Frossard, P.}
\newblock \bibinfo{title}{Generative modelling of structurally constrained graphs}.
\newblock \emph{\bibinfo{journal}{Advances in Neural Information Processing Systems}} \textbf{\bibinfo{volume}{37}}, \bibinfo{pages}{137218--137262} (\bibinfo{year}{2024}).

\bibitem{kengkanna2024enhancing}
\bibinfo{author}{Kengkanna, A.} \& \bibinfo{author}{Ohue, M.}
\newblock \bibinfo{title}{Enhancing property and activity prediction and interpretation using multiple molecular graph representations with mmgx}.
\newblock \emph{\bibinfo{journal}{Communications Chemistry}} \textbf{\bibinfo{volume}{7}}, \bibinfo{pages}{74} (\bibinfo{year}{2024}).

\bibitem{jin2018junction}
\bibinfo{author}{Jin, W.}, \bibinfo{author}{Barzilay, R.} \& \bibinfo{author}{Jaakkola, T.}
\newblock \bibinfo{title}{Junction tree variational autoencoder for molecular graph generation}.
\newblock In \emph{\bibinfo{booktitle}{International conference on machine learning}}, \bibinfo{pages}{2323--2332} (\bibinfo{organization}{PMLR}, \bibinfo{year}{2018}).

\bibitem{kong2022principal}
\bibinfo{author}{Kong, X.}, \bibinfo{author}{Huang, W.}, \bibinfo{author}{Tan, Z.} \& \bibinfo{author}{Liu, Y.}
\newblock \bibinfo{title}{Molecule generation by principal subgraph mining and assembling}.
\newblock \emph{\bibinfo{journal}{Advances in Neural Information Processing Systems}} \textbf{\bibinfo{volume}{35}}, \bibinfo{pages}{2550--2563} (\bibinfo{year}{2022}).

\bibitem{jin2020multi}
\bibinfo{author}{Jin, W.}, \bibinfo{author}{Barzilay, R.} \& \bibinfo{author}{Jaakkola, T.}
\newblock \bibinfo{title}{Multi-objective molecule generation using interpretable substructures}.
\newblock In \emph{\bibinfo{booktitle}{International conference on machine learning}}, \bibinfo{pages}{4849--4859} (\bibinfo{organization}{PMLR}, \bibinfo{year}{2020}).

\bibitem{stoehr2019disentangling}
\bibinfo{author}{Stoehr, N.}, \bibinfo{author}{Yilmaz, E.}, \bibinfo{author}{Brockschmidt, M.} \& \bibinfo{author}{Stuehmer, J.}
\newblock \bibinfo{title}{Disentangling interpretable generative parameters of random and real-world graphs}.
\newblock \emph{\bibinfo{journal}{arXiv preprint arXiv:1910.05639}}  (\bibinfo{year}{2019}).

\bibitem{guo2020interpretable}
\bibinfo{author}{Guo, X.} \emph{et~al.}
\newblock \bibinfo{title}{Interpretable deep graph generation with node-edge co-disentanglement}.
\newblock In \emph{\bibinfo{booktitle}{Proceedings of the 26th ACM SIGKDD international conference on knowledge discovery \& data mining}}, \bibinfo{pages}{1697--1707} (\bibinfo{year}{2020}).

\bibitem{blei2003latent}
\bibinfo{author}{Blei, D.~M.}, \bibinfo{author}{Ng, A.~Y.} \& \bibinfo{author}{Jordan, M.~I.}
\newblock \bibinfo{title}{Latent dirichlet allocation}.
\newblock \emph{\bibinfo{journal}{Journal of machine Learning research}} \textbf{\bibinfo{volume}{3}}, \bibinfo{pages}{993--1022} (\bibinfo{year}{2003}).

\bibitem{sohn2015learning}
\bibinfo{author}{Sohn, K.}, \bibinfo{author}{Lee, H.} \& \bibinfo{author}{Yan, X.}
\newblock \bibinfo{title}{Learning structured output representation using deep conditional generative models}.
\newblock \emph{\bibinfo{journal}{Advances in neural information processing systems}} \textbf{\bibinfo{volume}{28}} (\bibinfo{year}{2015}).

\bibitem{GraphVAE}
\bibinfo{author}{Simonovsky, M.} \& \bibinfo{author}{Komodakis, N.}
\newblock \bibinfo{title}{Graphvae: Towards generation of small graphs using variational autoencoders}.
\newblock \emph{\bibinfo{journal}{Le Centre pour la Communication Scientifique Directe - HAL - Diderot,Le Centre pour la Communication Scientifique Directe - HAL - Diderot}}  (\bibinfo{year}{2018}).

\bibitem{GraphVAE–MM}
\bibinfo{author}{Zahirnia, K.}, \bibinfo{author}{Schulte, O.}, \bibinfo{author}{Naddaf, P.} \& \bibinfo{author}{Li, K.}
\newblock \bibinfo{title}{Micro and macro level graph modeling for graph variational auto-encoders}  (\bibinfo{year}{2022}).

\bibitem{GraphRNN}
\bibinfo{author}{You, J.}, \bibinfo{author}{Ying, R.}, \bibinfo{author}{Ren, X.}, \bibinfo{author}{Hamilton, W.} \& \bibinfo{author}{Leskovec, J.}
\newblock \bibinfo{title}{Graphrnn: Generating realistic graphs with deep auto-regressive models}.
\newblock \emph{\bibinfo{journal}{International Conference on Machine Learning,International Conference on Machine Learning}}  (\bibinfo{year}{2018}).

\bibitem{GRAN}
\bibinfo{author}{Liao, R.} \emph{et~al.}
\newblock \bibinfo{title}{Efficient graph generation with graph recurrent attention networks}.
\newblock \emph{\bibinfo{journal}{Advances in neural information processing systems}} \textbf{\bibinfo{volume}{32}} (\bibinfo{year}{2019}).

\bibitem{BiGG}
\bibinfo{author}{Dai, H.}, \bibinfo{author}{Nazi, A.}, \bibinfo{author}{Li, Y.}, \bibinfo{author}{Dai, B.} \& \bibinfo{author}{Schuurmans, D.}
\newblock \bibinfo{title}{Scalable deep generative modeling for sparse graphs}.
\newblock \emph{\bibinfo{journal}{International Conference on Machine Learning,International Conference on Machine Learning}}  (\bibinfo{year}{2020}).

\bibitem{vignac2022digress}
\bibinfo{author}{Vignac, C.} \emph{et~al.}
\newblock \bibinfo{title}{Digress: Discrete denoising diffusion for graph generation}.
\newblock \emph{\bibinfo{journal}{arXiv preprint arXiv:2209.14734}}  (\bibinfo{year}{2022}).

\bibitem{golomb1996polyominoes}
\bibinfo{author}{Golomb, S.~W.}
\newblock \emph{\bibinfo{title}{Polyominoes: puzzles, patterns, problems, and packings}}, vol.~\bibinfo{volume}{16} (\bibinfo{publisher}{Princeton University Press}, \bibinfo{year}{1996}).

\bibitem{debnath1991structure}
\bibinfo{author}{Debnath, A.~K.}, \bibinfo{author}{Lopez~de Compadre, R.~L.}, \bibinfo{author}{Debnath, G.}, \bibinfo{author}{Shusterman, A.~J.} \& \bibinfo{author}{Hansch, C.}
\newblock \bibinfo{title}{Structure-activity relationship of mutagenic aromatic and heteroaromatic nitro compounds. correlation with molecular orbital energies and hydrophobicity}.
\newblock \emph{\bibinfo{journal}{Journal of medicinal chemistry}} \textbf{\bibinfo{volume}{34}}, \bibinfo{pages}{786--797} (\bibinfo{year}{1991}).

\bibitem{toivonen2003statistical}
\bibinfo{author}{Toivonen, H.}, \bibinfo{author}{Srinivasan, A.}, \bibinfo{author}{King, R.~D.}, \bibinfo{author}{Kramer, S.} \& \bibinfo{author}{Helma, C.}
\newblock \bibinfo{title}{Statistical evaluation of the predictive toxicology challenge 2000--2001}.
\newblock \emph{\bibinfo{journal}{Bioinformatics}} \textbf{\bibinfo{volume}{19}}, \bibinfo{pages}{1183--1193} (\bibinfo{year}{2003}).

\bibitem{xu2018powerful}
\bibinfo{author}{Xu, K.}, \bibinfo{author}{Hu, W.}, \bibinfo{author}{Leskovec, J.} \& \bibinfo{author}{Jegelka, S.}
\newblock \bibinfo{title}{How powerful are graph neural networks?}
\newblock \emph{\bibinfo{journal}{arXiv preprint arXiv:1810.00826}}  (\bibinfo{year}{2018}).

\bibitem{hu2020open}
\bibinfo{author}{Hu, W.} \emph{et~al.}
\newblock \bibinfo{title}{Open graph benchmark: Datasets for machine learning on graphs}.
\newblock \emph{\bibinfo{journal}{Advances in neural information processing systems}} \textbf{\bibinfo{volume}{33}}, \bibinfo{pages}{22118--22133} (\bibinfo{year}{2020}).

\bibitem{thompson2022evaluation}
\bibinfo{author}{Thompson, R.}, \bibinfo{author}{Knyazev, B.}, \bibinfo{author}{Ghalebi, E.}, \bibinfo{author}{Kim, J.} \& \bibinfo{author}{Taylor, G.~W.}
\newblock \bibinfo{title}{On evaluation metrics for graph generative models}.
\newblock \emph{\bibinfo{journal}{arXiv preprint arXiv:2201.09871}}  (\bibinfo{year}{2022}).

\bibitem{OBray_Horn_Rieck_Borgwardt_2021}
\bibinfo{author}{O’Bray, L.}, \bibinfo{author}{Horn, M.}, \bibinfo{author}{Rieck, B.} \& \bibinfo{author}{Borgwardt, K.}
\newblock \bibinfo{title}{Evaluation metrics for graph generative models: Problems, pitfalls, and practical solutions.}
\newblock \emph{\bibinfo{journal}{Learning,Learning}}  (\bibinfo{year}{2021}).

\bibitem{gonzalez2001application}
\bibinfo{author}{Gonzalez, J.}, \bibinfo{author}{Holder, L.} \& \bibinfo{author}{Cook, D.~J.}
\newblock \bibinfo{title}{Application of graph-based concept learning to the predictive toxicology domain}.
\newblock In \emph{\bibinfo{booktitle}{Proceedings of the Predictive Toxicology Challenge Workshop}} (\bibinfo{year}{2001}).

\bibitem{swamidass2005kernels}
\bibinfo{author}{Swamidass, S.~J.} \emph{et~al.}
\newblock \bibinfo{title}{Kernels for small molecules and the prediction of mutagenicity, toxicity and anti-cancer activity}.
\newblock In \emph{\bibinfo{booktitle}{ISMB (Supplement of Bioinformatics)}}, \bibinfo{pages}{359--368} (\bibinfo{year}{2005}).

\bibitem{blinova2003toxicology}
\bibinfo{author}{Blinova, V.}, \bibinfo{author}{Dobrynin, D.}, \bibinfo{author}{Finn, V.~K.}, \bibinfo{author}{Kuznetsov, S.~O.} \& \bibinfo{author}{Pankratova, E.}
\newblock \bibinfo{title}{Toxicology analysis by means of the jsm-method}.
\newblock \emph{\bibinfo{journal}{Bioinformatics}} \textbf{\bibinfo{volume}{19}}, \bibinfo{pages}{1201--1207} (\bibinfo{year}{2003}).

\bibitem{de2018molgan}
\bibinfo{author}{De~Cao, N.} \& \bibinfo{author}{Kipf, T.}
\newblock \bibinfo{title}{Molgan: An implicit generative model for small molecular graphs}.
\newblock \emph{\bibinfo{journal}{arXiv preprint arXiv:1805.11973}}  (\bibinfo{year}{2018}).

\bibitem{wang2023fairness}
\bibinfo{author}{Wang, Z.}, \bibinfo{author}{Wallace, C.}, \bibinfo{author}{Bifet, A.}, \bibinfo{author}{Yao, X.} \& \bibinfo{author}{Zhang, W.}
\newblock \bibinfo{title}{Fairness-aware graph generative adversarial networks}.
\newblock In \emph{\bibinfo{booktitle}{Joint European Conference on Machine Learning and Knowledge Discovery in Databases}}, \bibinfo{pages}{259--275} (\bibinfo{organization}{Springer}, \bibinfo{year}{2023}).

\bibitem{kong2023autoregressive}
\bibinfo{author}{Kong, L.} \emph{et~al.}
\newblock \bibinfo{title}{Autoregressive diffusion model for graph generation}.
\newblock In \emph{\bibinfo{booktitle}{International conference on machine learning}}, \bibinfo{pages}{17391--17408} (\bibinfo{organization}{PMLR}, \bibinfo{year}{2023}).

\bibitem{bu2023let}
\bibinfo{author}{Bu, J.}, \bibinfo{author}{Mehrab, K.~S.} \& \bibinfo{author}{Karpatne, A.}
\newblock \bibinfo{title}{Let there be order: Rethinking ordering in autoregressive graph generation}.
\newblock \emph{\bibinfo{journal}{arXiv preprint arXiv:2305.15562}}  (\bibinfo{year}{2023}).

\bibitem{luo2021graphdf}
\bibinfo{author}{Luo, Y.}, \bibinfo{author}{Yan, K.} \& \bibinfo{author}{Ji, S.}
\newblock \bibinfo{title}{Graphdf: A discrete flow model for molecular graph generation}.
\newblock In \emph{\bibinfo{booktitle}{International conference on machine learning}}, \bibinfo{pages}{7192--7203} (\bibinfo{organization}{PMLR}, \bibinfo{year}{2021}).

\bibitem{kuznetsov2021molgrow}
\bibinfo{author}{Kuznetsov, M.} \& \bibinfo{author}{Polykovskiy, D.}
\newblock \bibinfo{title}{Molgrow: A graph normalizing flow for hierarchical molecular generation}.
\newblock In \emph{\bibinfo{booktitle}{Proceedings of the AAAI Conference on Artificial Intelligence}}, vol.~\bibinfo{volume}{35}, \bibinfo{pages}{8226--8234} (\bibinfo{year}{2021}).

\bibitem{cho2024multi}
\bibinfo{author}{Cho, H.}, \bibinfo{author}{Jeong, M.}, \bibinfo{author}{Jeon, S.}, \bibinfo{author}{Ahn, S.} \& \bibinfo{author}{Kim, W.~H.}
\newblock \bibinfo{title}{Multi-resolution spectral coherence for graph generation with score-based diffusion}.
\newblock \emph{\bibinfo{journal}{Advances in Neural Information Processing Systems}} \textbf{\bibinfo{volume}{36}} (\bibinfo{year}{2024}).

\bibitem{minello2024graph}
\bibinfo{author}{Minello, G.}, \bibinfo{author}{Bicciato, A.}, \bibinfo{author}{Rossi, L.}, \bibinfo{author}{Torsello, A.} \& \bibinfo{author}{Cosmo, L.}
\newblock \bibinfo{title}{Graph generation via spectral diffusion}.
\newblock \emph{\bibinfo{journal}{arXiv preprint arXiv:2402.18974}}  (\bibinfo{year}{2024}).

\bibitem{zang2020moflow}
\bibinfo{author}{Zang, C.} \& \bibinfo{author}{Wang, F.}
\newblock \bibinfo{title}{Moflow: an invertible flow model for generating molecular graphs}.
\newblock In \emph{\bibinfo{booktitle}{Proceedings of the 26th ACM SIGKDD international conference on knowledge discovery \& data mining}}, \bibinfo{pages}{617--626} (\bibinfo{year}{2020}).

\bibitem{brockschmidt2018generative}
\bibinfo{author}{Brockschmidt, M.}, \bibinfo{author}{Allamanis, M.}, \bibinfo{author}{Gaunt, A.~L.} \& \bibinfo{author}{Polozov, O.}
\newblock \bibinfo{title}{Generative code modeling with graphs}.
\newblock \emph{\bibinfo{journal}{arXiv preprint arXiv:1805.08490}}  (\bibinfo{year}{2018}).

\bibitem{dai2018syntax}
\bibinfo{author}{Dai, H.}, \bibinfo{author}{Tian, Y.}, \bibinfo{author}{Dai, B.}, \bibinfo{author}{Skiena, S.} \& \bibinfo{author}{Song, L.}
\newblock \bibinfo{title}{Syntax-directed variational autoencoder for structured data}.
\newblock \emph{\bibinfo{journal}{arXiv preprint arXiv:1802.08786}}  (\bibinfo{year}{2018}).

\bibitem{li2020dirichlet}
\bibinfo{author}{Li, J.} \emph{et~al.}
\newblock \bibinfo{title}{Dirichlet graph variational autoencoder}.
\newblock \emph{\bibinfo{journal}{Advances in Neural Information Processing Systems}} \textbf{\bibinfo{volume}{33}}, \bibinfo{pages}{5274--5283} (\bibinfo{year}{2020}).

\bibitem{du2022interpretable}
\bibinfo{author}{Du, Y.}, \bibinfo{author}{Guo, X.}, \bibinfo{author}{Shehu, A.} \& \bibinfo{author}{Zhao, L.}
\newblock \bibinfo{title}{Interpretable molecular graph generation via monotonic constraints}.
\newblock In \emph{\bibinfo{booktitle}{Proceedings of the 2022 SIAM International Conference on Data Mining (SDM)}}, \bibinfo{pages}{73--81} (\bibinfo{organization}{SIAM}, \bibinfo{year}{2022}).

\bibitem{hofmann2001unsupervised}
\bibinfo{author}{Hofmann, T.}
\newblock \bibinfo{title}{Unsupervised learning by probabilistic latent semantic analysis}.
\newblock \emph{\bibinfo{journal}{Machine learning}} \textbf{\bibinfo{volume}{42}}, \bibinfo{pages}{177--196} (\bibinfo{year}{2001}).

\bibitem{lee2000algorithms}
\bibinfo{author}{Lee, D.} \& \bibinfo{author}{Seung, H.~S.}
\newblock \bibinfo{title}{Algorithms for non-negative matrix factorization}.
\newblock \emph{\bibinfo{journal}{Advances in neural information processing systems}} \textbf{\bibinfo{volume}{13}} (\bibinfo{year}{2000}).

\bibitem{miao2016neural}
\bibinfo{author}{Miao, Y.}, \bibinfo{author}{Yu, L.} \& \bibinfo{author}{Blunsom, P.}
\newblock \bibinfo{title}{Neural variational inference for text processing}.
\newblock In \emph{\bibinfo{booktitle}{International conference on machine learning}}, \bibinfo{pages}{1727--1736} (\bibinfo{organization}{PMLR}, \bibinfo{year}{2016}).

\bibitem{miao2017discovering}
\bibinfo{author}{Miao, Y.}, \bibinfo{author}{Grefenstette, E.} \& \bibinfo{author}{Blunsom, P.}
\newblock \bibinfo{title}{Discovering discrete latent topics with neural variational inference}.
\newblock In \emph{\bibinfo{booktitle}{International conference on machine learning}}, \bibinfo{pages}{2410--2419} (\bibinfo{organization}{PMLR}, \bibinfo{year}{2017}).

\bibitem{bianchi2020pre}
\bibinfo{author}{Bianchi, F.}, \bibinfo{author}{Terragni, S.} \& \bibinfo{author}{Hovy, D.}
\newblock \bibinfo{title}{Pre-training is a hot topic: Contextualized document embeddings improve topic coherence}.
\newblock \emph{\bibinfo{journal}{arXiv preprint arXiv:2004.03974}}  (\bibinfo{year}{2020}).

\bibitem{long2020graph}
\bibinfo{author}{Long, Q.}, \bibinfo{author}{Jin, Y.}, \bibinfo{author}{Song, G.}, \bibinfo{author}{Li, Y.} \& \bibinfo{author}{Lin, W.}
\newblock \bibinfo{title}{Graph structural-topic neural network}.
\newblock In \emph{\bibinfo{booktitle}{Proceedings of the 26th ACM SIGKDD international conference on knowledge discovery \& data mining}}, \bibinfo{pages}{1065--1073} (\bibinfo{year}{2020}).

\bibitem{xie2021graph}
\bibinfo{author}{Xie, Q.}, \bibinfo{author}{Zhu, Y.}, \bibinfo{author}{Huang, J.}, \bibinfo{author}{Du, P.} \& \bibinfo{author}{Nie, J.-Y.}
\newblock \bibinfo{title}{Graph neural collaborative topic model for citation recommendation}.
\newblock \emph{\bibinfo{journal}{ACM Transactions on Information Systems (TOIS)}} \textbf{\bibinfo{volume}{40}}, \bibinfo{pages}{1--30} (\bibinfo{year}{2021}).

\bibitem{sonderby2016ladder}
\bibinfo{author}{S{\o}nderby, C.~K.}, \bibinfo{author}{Raiko, T.}, \bibinfo{author}{Maal{\o}e, L.}, \bibinfo{author}{S{\o}nderby, S.~K.} \& \bibinfo{author}{Winther, O.}
\newblock \bibinfo{title}{Ladder variational autoencoders}.
\newblock \emph{\bibinfo{journal}{Advances in neural information processing systems}} \textbf{\bibinfo{volume}{29}} (\bibinfo{year}{2016}).

\bibitem{kingma2013auto}
\bibinfo{author}{Kingma, D.~P.} \& \bibinfo{author}{Welling, M.}
\newblock \bibinfo{title}{Auto-encoding variational bayes}.
\newblock \emph{\bibinfo{journal}{arXiv preprint arXiv:1312.6114}}  (\bibinfo{year}{2013}).

\bibitem{blei2017variational}
\bibinfo{author}{Blei, D.~M.}, \bibinfo{author}{Kucukelbir, A.} \& \bibinfo{author}{McAuliffe, J.~D.}
\newblock \bibinfo{title}{Variational inference: A review for statisticians}.
\newblock \emph{\bibinfo{journal}{Journal of the American statistical Association}} \textbf{\bibinfo{volume}{112}}, \bibinfo{pages}{859--877} (\bibinfo{year}{2017}).

\bibitem{ba2016layer}
\bibinfo{author}{Ba, J.~L.}, \bibinfo{author}{Kiros, J.~R.} \& \bibinfo{author}{Hinton, G.~E.}
\newblock \bibinfo{title}{Layer normalization}.
\newblock \emph{\bibinfo{journal}{arXiv preprint arXiv:1607.06450}}  (\bibinfo{year}{2016}).

\end{thebibliography}

\clearpage

\begin{appendices}

\section{Appendix}

\subsection{Setup and Hyperparameters}

The IToG framework employs three encoders, each utilizing a two-layer Graph Convolutional Network (GCN) followed by a graph-level output formulation (summing the nodes' representations). The substructure decoder is a three-layer fully connected neural network that directly maps the substructure representation \(z^{w}\) to a probabilistic adjacency matrix. Each layer incorporates Layer Normalization~\cite{ba2016layer} and the LeakyReLU activation function.
The mapping encoder leverages a transformer model to process the combined substructures \(S\) and global information \(z^g\), employing multiple heads and layers. The mapping decoder is a two-layer fully connected network that converts the encoded mapping representations into the final mapping matrix format.
Training is performed using the Adam optimizer with a learning rate of 0.0003 for 20,000 epochs, except for the ogbg-molbbbp dataset, which is trained for 10,000 epochs. Table~\ref{tab:hype} lists the hyperparameters \(\alpha_{\text{MM}}\), \(\beta_{\text{KL}}\), \(\gamma_{\text{KL}}\), \(\delta_{\text{KL}}\), \(\omega_{\text{ortho}}\), the number of substructures \(W\), the number of topics \(K\), and the max node number of substructure $n$. These hyperparameters are selected based on validation set performance. The code for all models is run on the same system, an AMD EPYC 7543 32-Core Processor 3.67 GHz and NVIDIA RTX A6000 GPUs.

\begin{table}[h]
\centering
\caption{Hyperparameters for each dataset used in graph generation task.}
\begin{tabular}{ccccccccc}
\hline
Dataset      & \(\alpha_{\text{MM}}\)  & \(\beta_{\text{KL}}\)   & \(\delta_{\text{KL}}\)  & \(\omega_{\text{ortho}}\)& \(\gamma_{\text{KL}}\)  & $K$  & $W$  & $n$ \\ \hline
MUTAG        & 80  & 90 & 15  & 10  & 10 & 5 & 20 & 8      \\
Lobster      & 20  & 10 & 0.5 & 1.5 & 1  & 5  & 20 & 10     \\
PTC          & 10  & 1  & 1   & 1   & 1  & 5  & 20 & 20     \\
Ogbg-molbbbp & 200 & 50 & 5   & 5   & 4  & 5  & 20 & 20     \\ \hline
\end{tabular}
\label{tab:hype}
\end{table}

\subsection{Qualitative Evaluation Results}
\label{appendix_visualization}
\begin{figure}[h]
  \centering
  \begin{minipage}{0.06\textwidth}
    \centering
    \rotatebox{90}{\small Test}
  \end{minipage}
  \begin{minipage}{0.11\textwidth}
    \includegraphics[width=\linewidth]{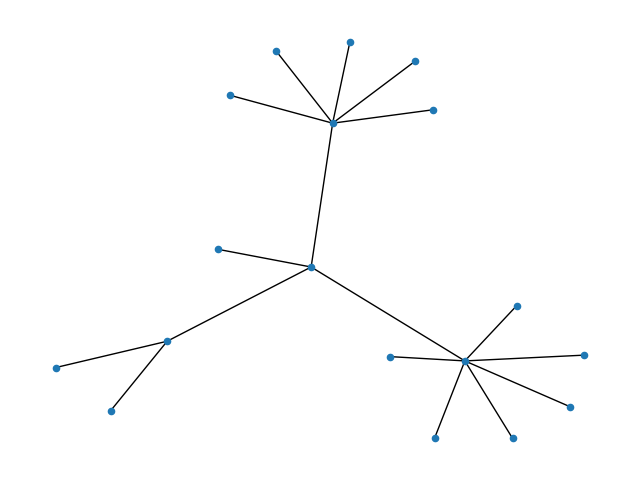}
  \end{minipage}
  \begin{minipage}{0.11\textwidth}
    \includegraphics[width=\linewidth]{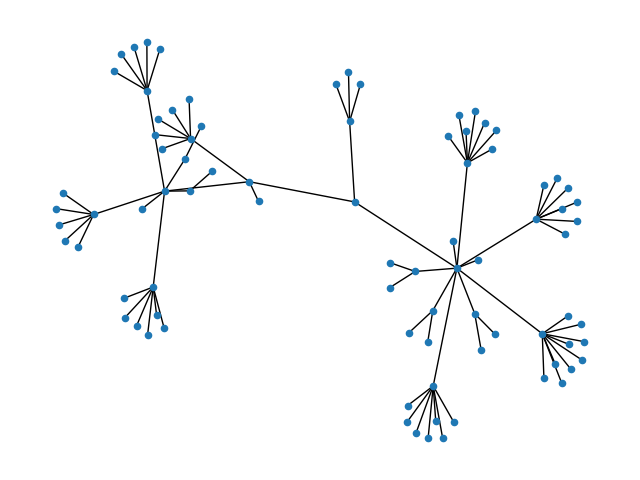}
  \end{minipage}
  \begin{minipage}{0.11\textwidth}
    \includegraphics[width=\linewidth]{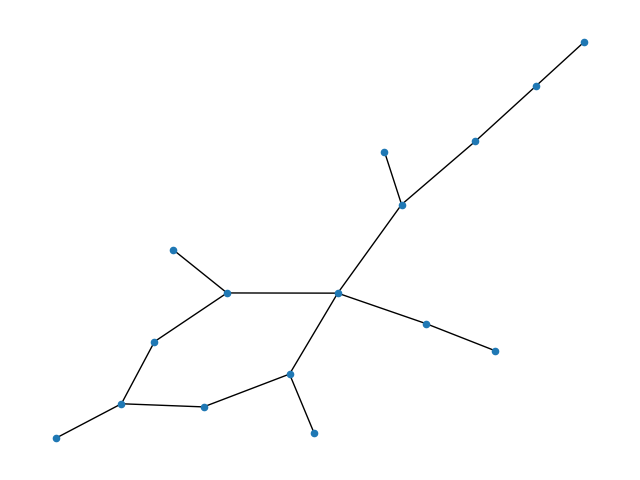}
  \end{minipage}
  \begin{minipage}{0.11\textwidth}
    \includegraphics[width=\linewidth]{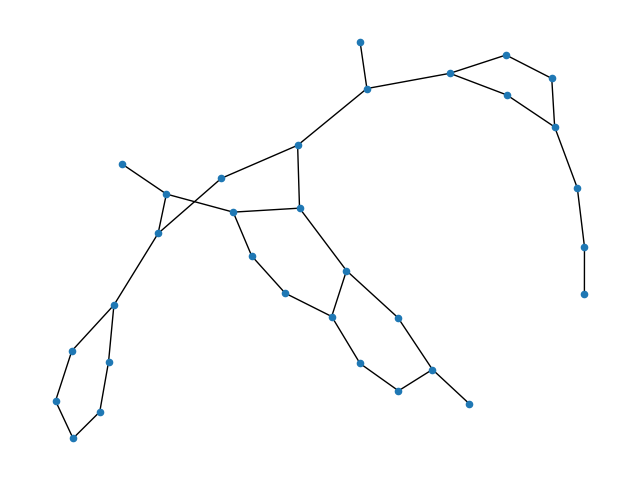}
  \end{minipage}
  \begin{minipage}{0.11\textwidth}
    \includegraphics[width=\linewidth]{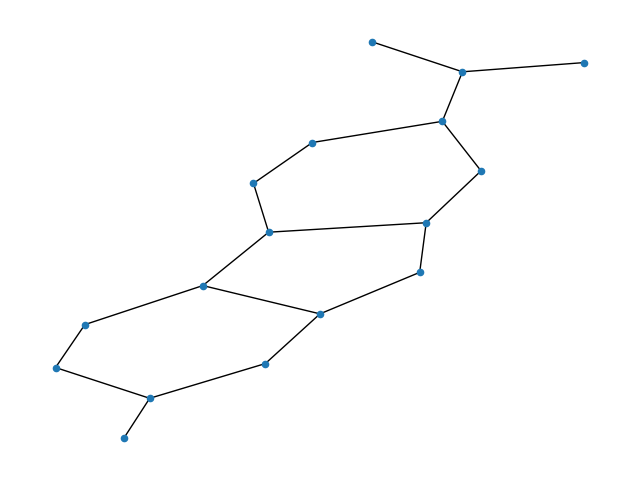}
  \end{minipage}
  \begin{minipage}{0.11\textwidth}
    \includegraphics[width=\linewidth]{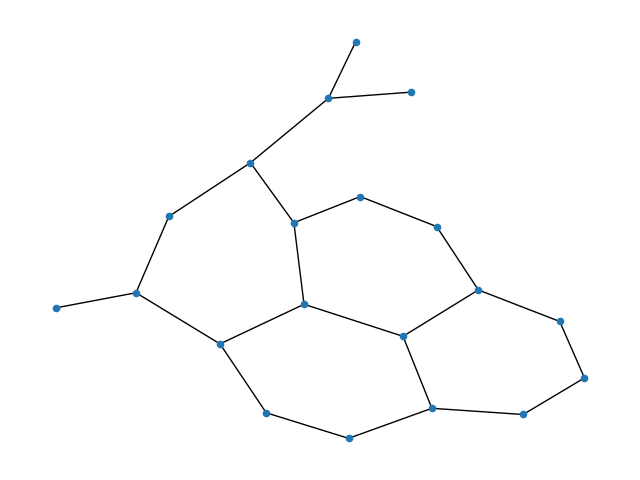}
  \end{minipage}
  \begin{minipage}{0.11\textwidth}
    \includegraphics[width=\linewidth]{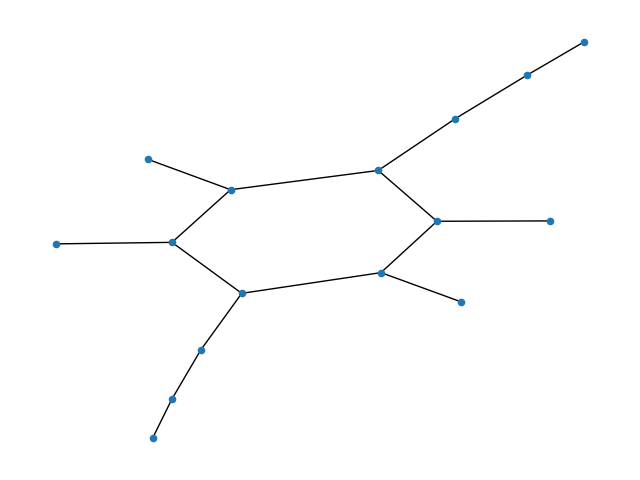}
  \end{minipage}
  \begin{minipage}{0.11\textwidth}
    \includegraphics[width=\linewidth]{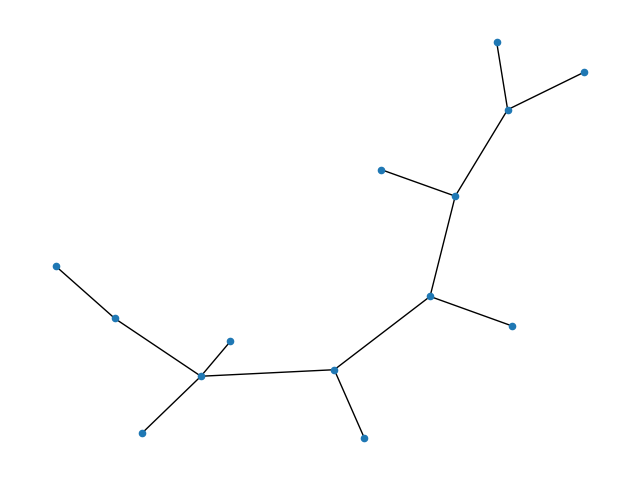}
  \end{minipage}

  \begin{minipage}{0.06\textwidth}
    \centering
    \rotatebox{90}{\small GraphRNN}
  \end{minipage}
  \begin{minipage}{0.11\textwidth}
    \includegraphics[width=\linewidth]{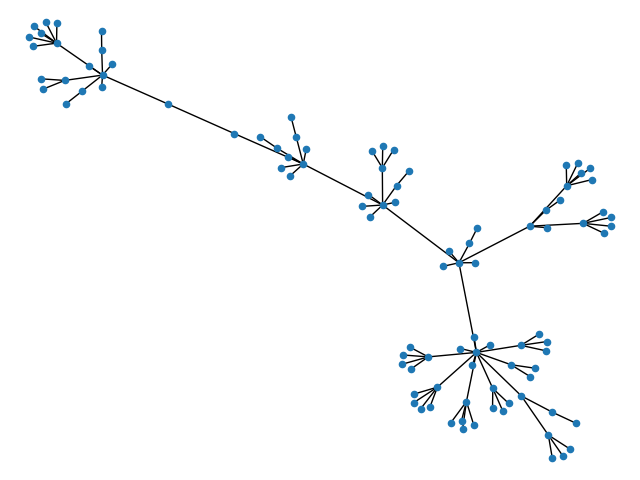}
  \end{minipage}
  \begin{minipage}{0.11\textwidth}
    \includegraphics[width=\linewidth]{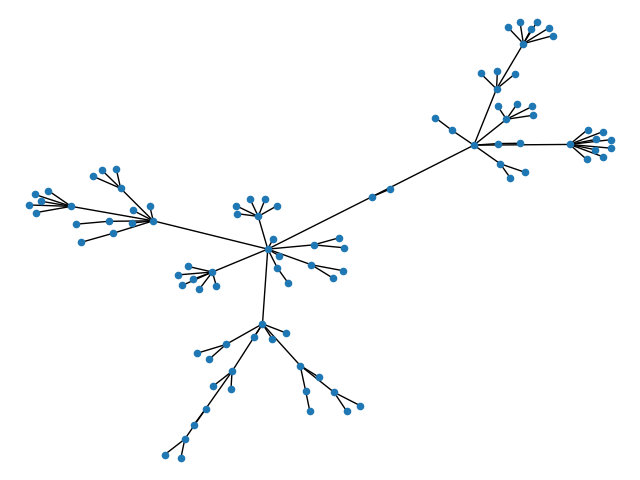}
  \end{minipage}
  \begin{minipage}{0.11\textwidth}
    \includegraphics[width=\linewidth]{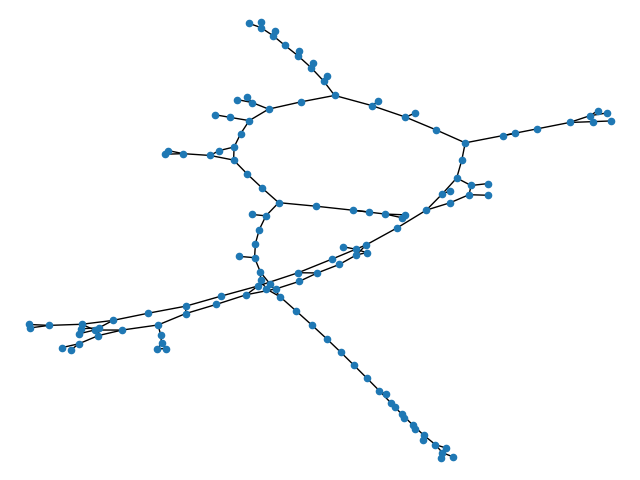}
  \end{minipage}
  \begin{minipage}{0.11\textwidth}
    \includegraphics[width=\linewidth]{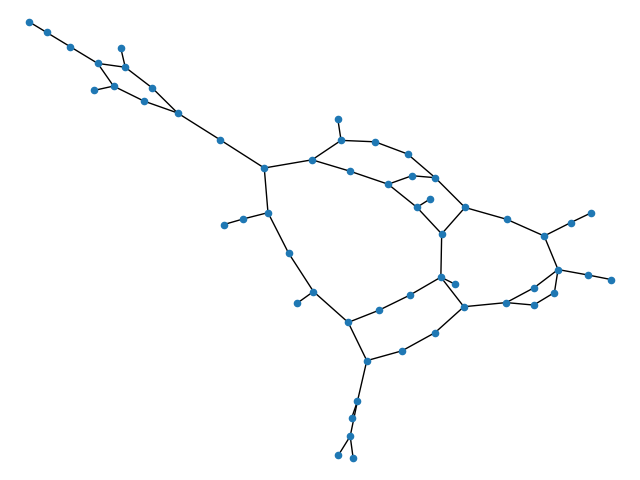}
  \end{minipage}
  \begin{minipage}{0.11\textwidth}
    \includegraphics[width=\linewidth]{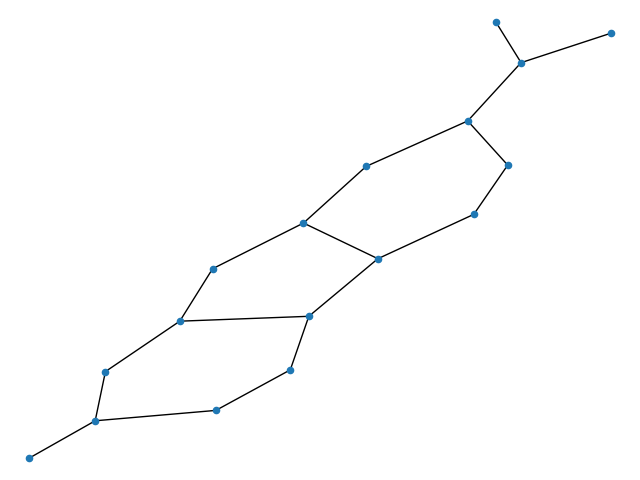}
  \end{minipage}
  \begin{minipage}{0.11\textwidth}
    \includegraphics[width=\linewidth]{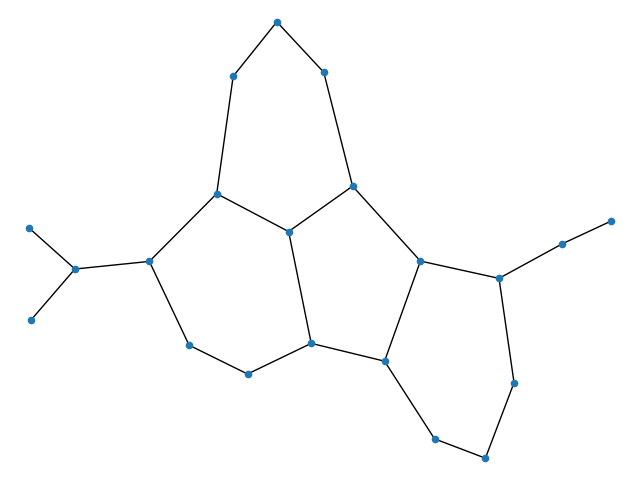}
  \end{minipage}
  \begin{minipage}{0.11\textwidth}
    \includegraphics[width=\linewidth]{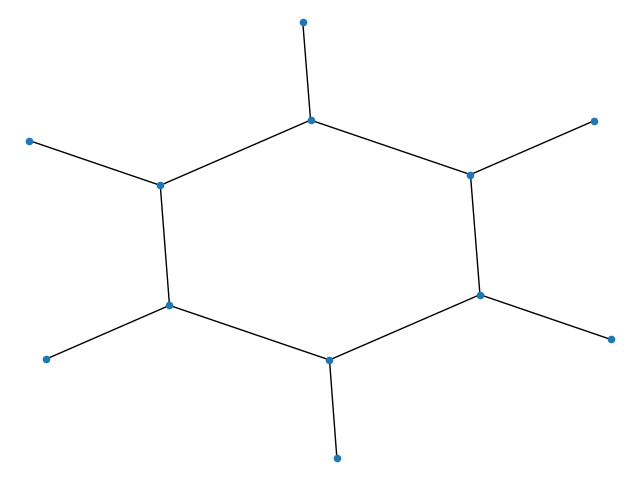}
  \end{minipage}
  \begin{minipage}{0.11\textwidth}
    \includegraphics[width=\linewidth]{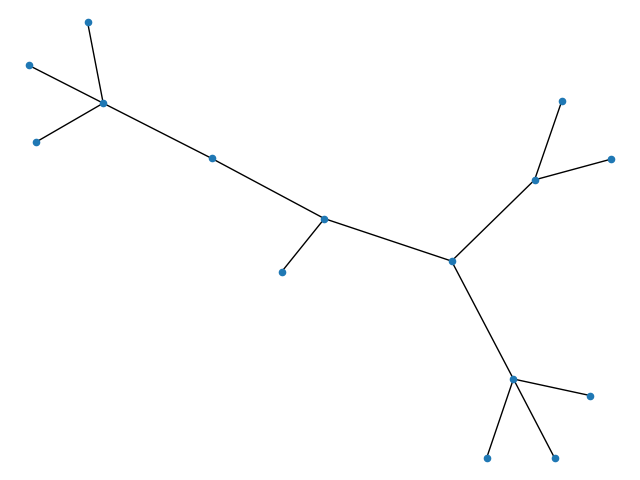}
  \end{minipage}

  \begin{minipage}{0.06\textwidth}
    \centering
    \rotatebox{90}{\small BIGG}
  \end{minipage}
  \begin{minipage}{0.11\textwidth}
    \includegraphics[width=\linewidth]{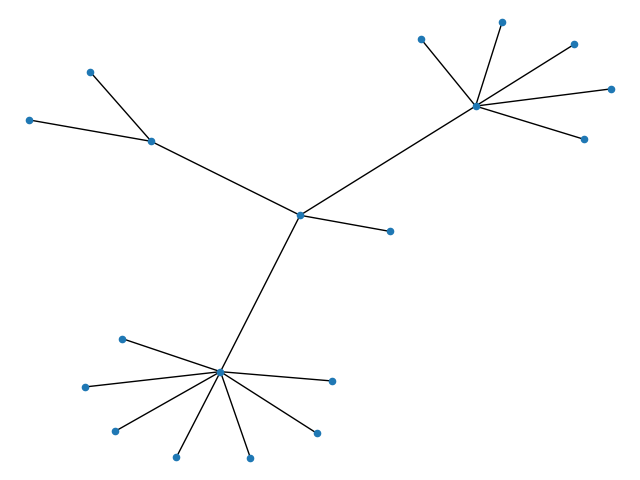}
  \end{minipage}
  \begin{minipage}{0.11\textwidth}
    \includegraphics[width=\linewidth]{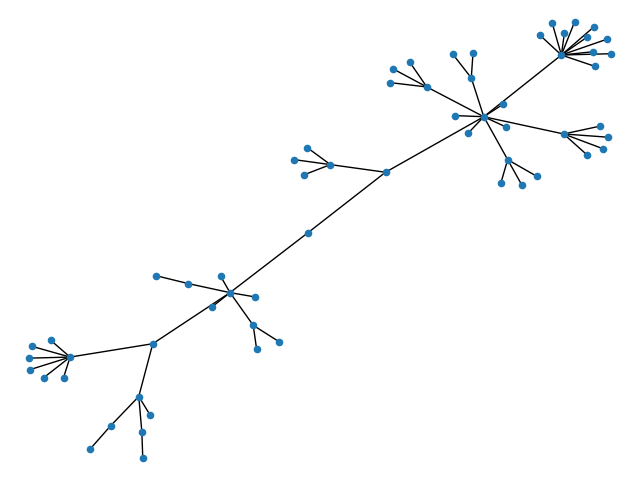}
  \end{minipage}
  \begin{minipage}{0.11\textwidth}
    \includegraphics[width=\linewidth]{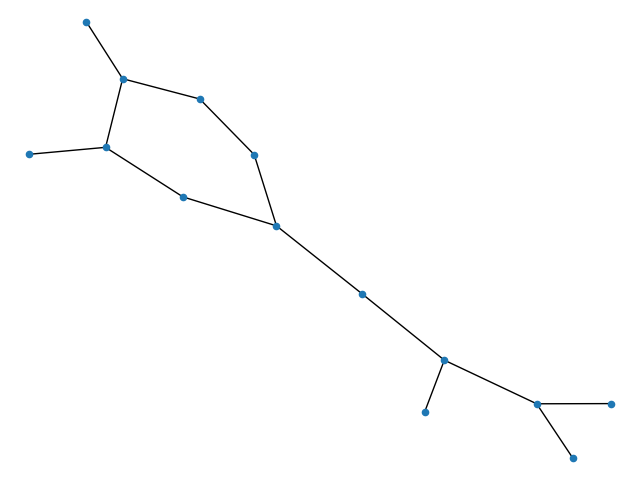}
  \end{minipage}
  \begin{minipage}{0.11\textwidth}
    \includegraphics[width=\linewidth]{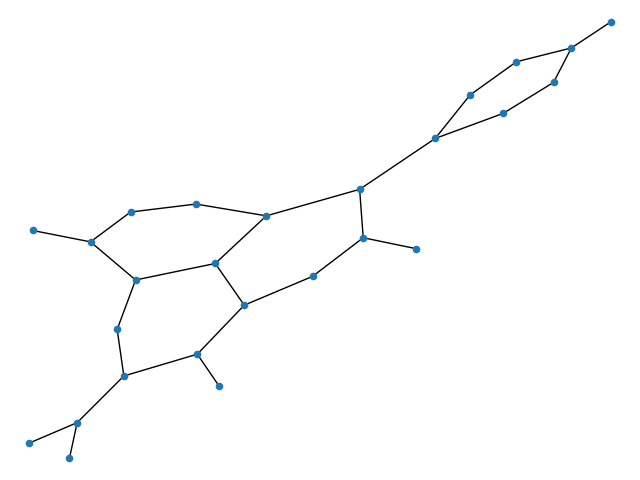}
  \end{minipage}
  \begin{minipage}{0.11\textwidth}
    \includegraphics[width=\linewidth]{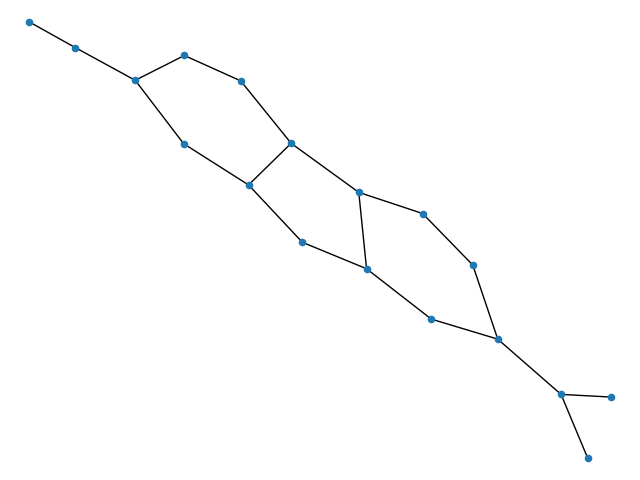}
  \end{minipage}
  \begin{minipage}{0.11\textwidth}
    \includegraphics[width=\linewidth]{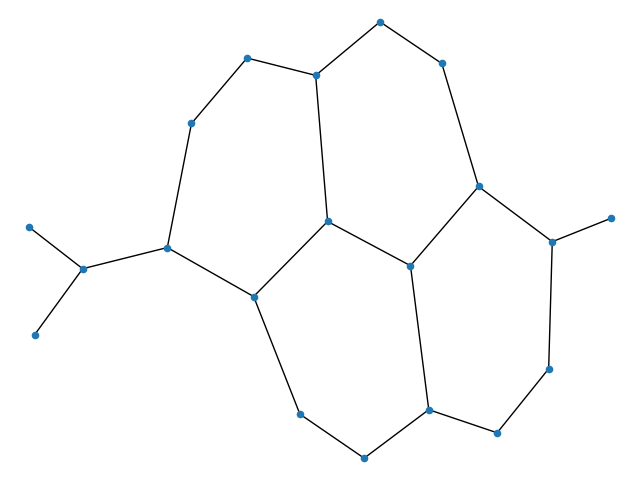}
  \end{minipage}
  \begin{minipage}{0.11\textwidth}
    \includegraphics[width=\linewidth]{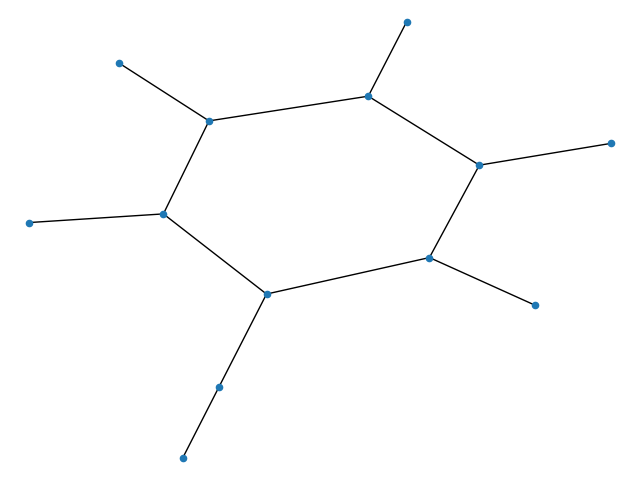}
  \end{minipage}
  \begin{minipage}{0.11\textwidth}
    \includegraphics[width=\linewidth]{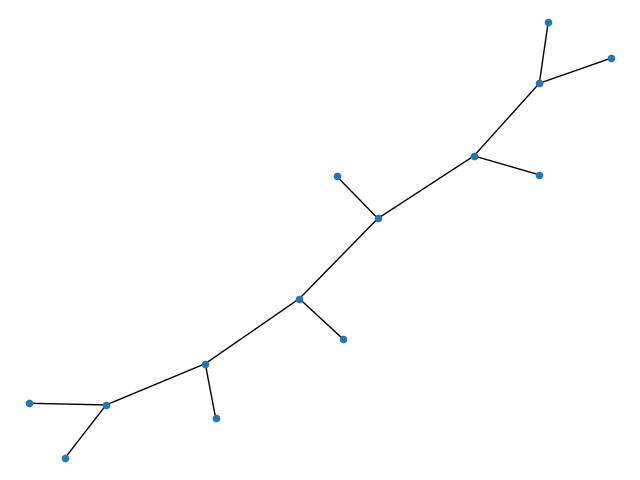}
  \end{minipage}

  \begin{minipage}{0.06\textwidth}
    \centering
    \rotatebox{90}{\small GraphVAE-MM}
  \end{minipage}
  \begin{minipage}{0.11\textwidth}
    \includegraphics[width=\linewidth]{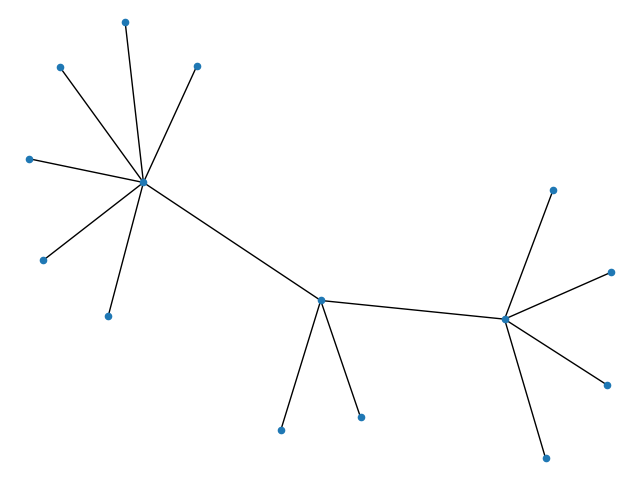}
  \end{minipage}
  \begin{minipage}{0.11\textwidth}
    \includegraphics[width=\linewidth]{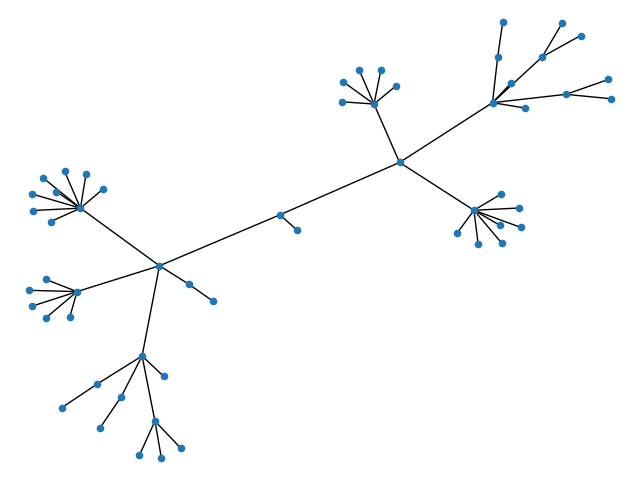}
  \end{minipage}
  \begin{minipage}{0.11\textwidth}
    \includegraphics[width=\linewidth]{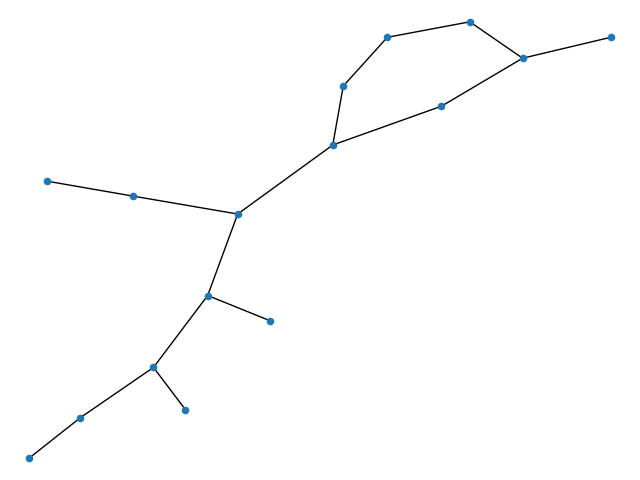}
  \end{minipage}
  \begin{minipage}{0.11\textwidth}
    \includegraphics[width=\linewidth]{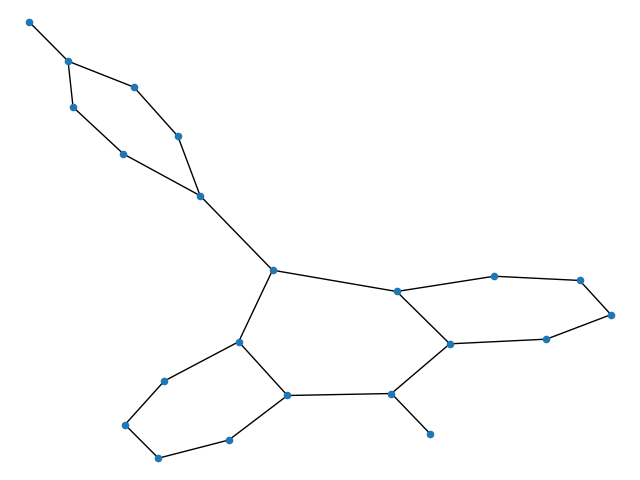}
  \end{minipage}
  \begin{minipage}{0.11\textwidth}
    \includegraphics[width=\linewidth]{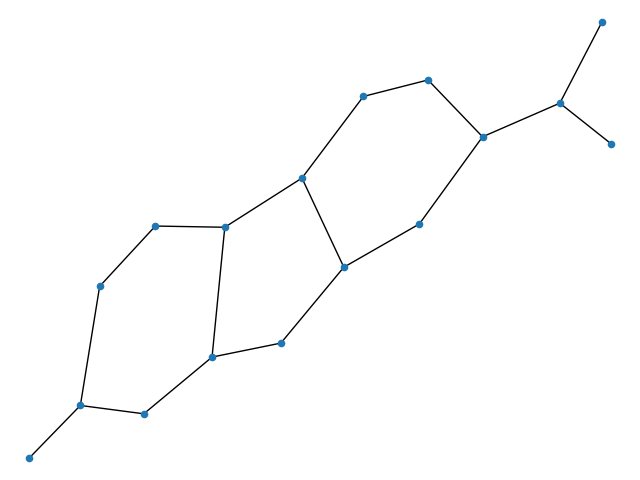}
  \end{minipage}
  \begin{minipage}{0.11\textwidth}
    \includegraphics[width=\linewidth]{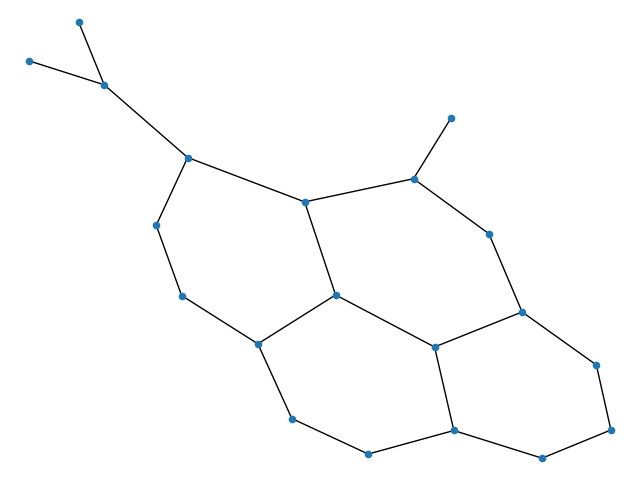}
  \end{minipage}
  \begin{minipage}{0.11\textwidth}
    \includegraphics[width=\linewidth]{picture/bigg_ptc1.png}
  \end{minipage}
  \begin{minipage}{0.11\textwidth}
    \includegraphics[width=\linewidth]{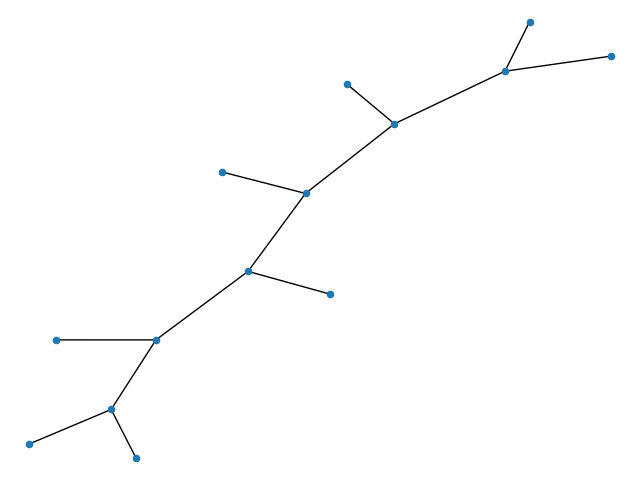}
  \end{minipage}

  \begin{minipage}{0.06\textwidth}
    \centering
    \rotatebox{90}{\small NGTM}
  \end{minipage}
  \begin{minipage}{0.11\textwidth}
    \includegraphics[width=\linewidth]{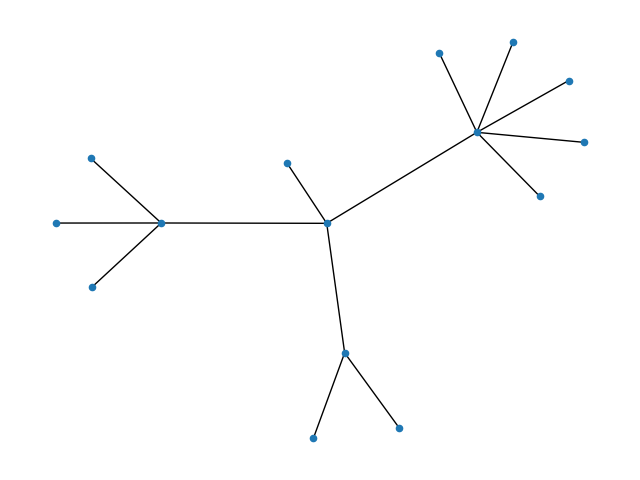}
  \end{minipage}
  \begin{minipage}{0.11\textwidth}
    \includegraphics[width=\linewidth]{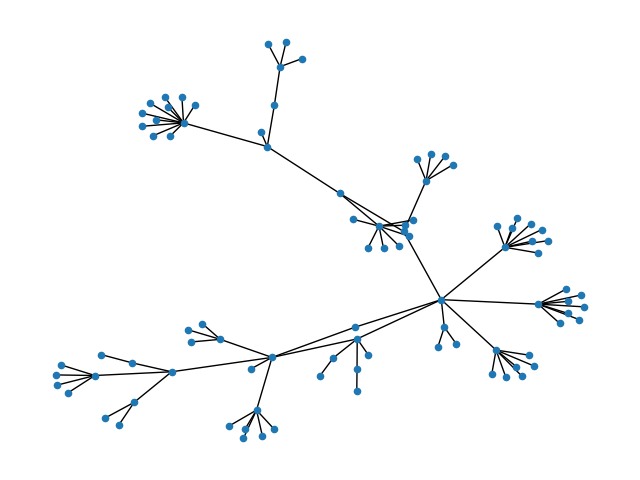}
  \end{minipage}
  \begin{minipage}{0.11\textwidth}
    \includegraphics[width=\linewidth]{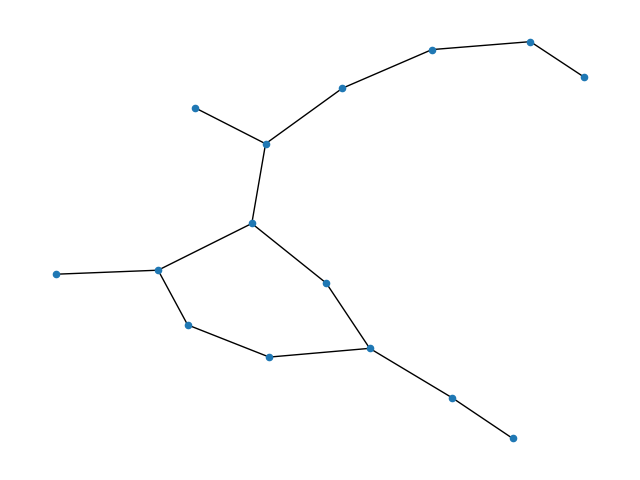}
  \end{minipage}
  \begin{minipage}{0.11\textwidth}
    \includegraphics[width=\linewidth]{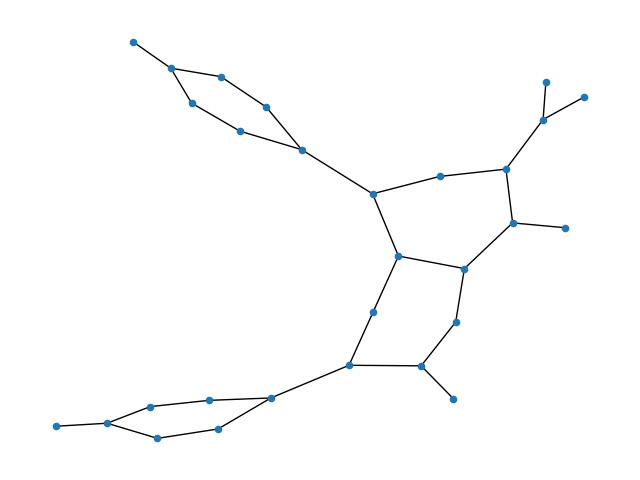}
  \end{minipage}
  \begin{minipage}{0.11\textwidth}
    \includegraphics[width=\linewidth]{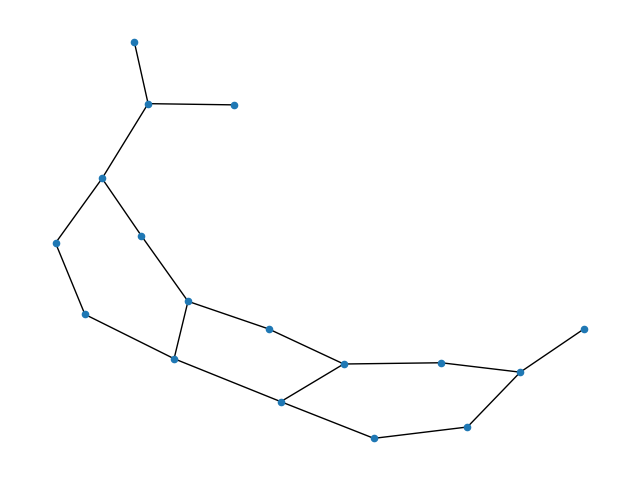}
  \end{minipage}
  \begin{minipage}{0.11\textwidth}
    \includegraphics[width=\linewidth]{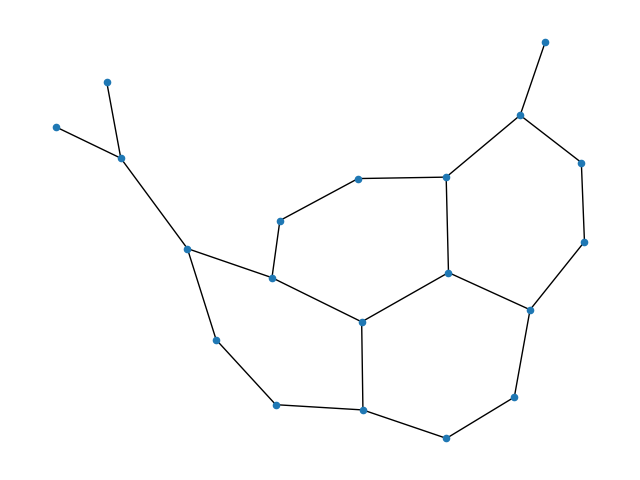}
  \end{minipage}
  \begin{minipage}{0.11\textwidth}
    \includegraphics[width=\linewidth]{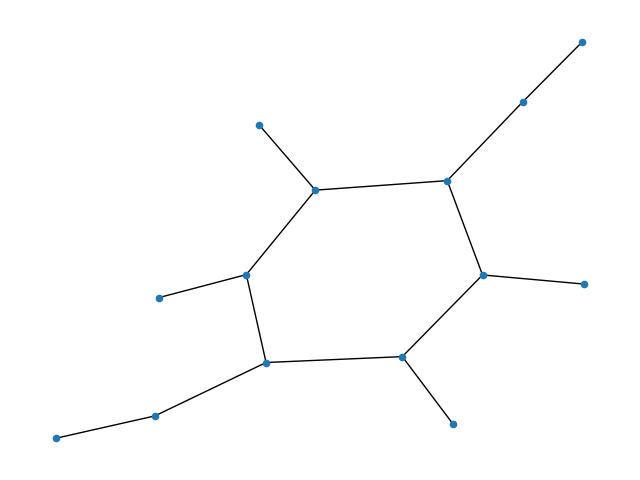}
  \end{minipage}
  \begin{minipage}{0.11\textwidth}
    \includegraphics[width=\linewidth]{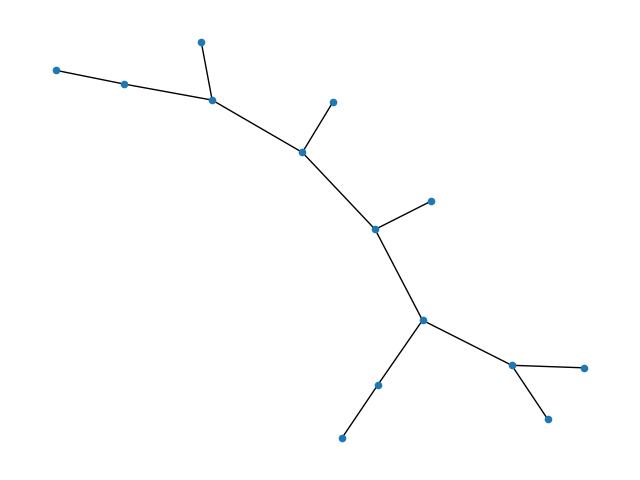}
  \end{minipage}

  \vspace{0.1cm}
  \begin{minipage}{0.06\textwidth}
  \end{minipage}
  \begin{minipage}{0.22\textwidth}
    \centering
    \textbf{\small Lobster}
  \end{minipage}
  \begin{minipage}{0.22\textwidth}
    \centering
    \textbf{\small Ogbg-molbbbp}
  \end{minipage}
  \begin{minipage}{0.22\textwidth}
    \centering
    \textbf{\small MUTAG}
  \end{minipage}
  \begin{minipage}{0.22\textwidth}
    \centering
    \textbf{\small PTC}
  \end{minipage}
  \vspace{-3pt}
  \caption{Visualization of generated graphs}
  \vspace{-3pt}
  \label{fig:Visualizations}
\end{figure}

Figure~\ref{fig:Visualizations} shows representative graphs generated by different models across four datasets, alongside randomly selected test samples. For each model, 20 samples are generated, and visually representative examples are shown for comparison.
On the Lobster dataset, NGTM is able to generate tree-like structures that resemble those in the test set, capturing the overall branching pattern. For Ogbg-molbbbp and MUTAG, NGTM produces molecular graphs that preserve common structural elements such as rings and chain fragments. Compared to baseline models, NGTM tends to generate graphs with clearer local patterns and fewer disconnected or redundant components.

\subsection{Extended Results: Effects of Topic Manipulation on Graph Properties}
\begin{figure}[ht]
  \centering
  \subfigure[T2-Clustering]{
      \includegraphics[width=0.42\textwidth]{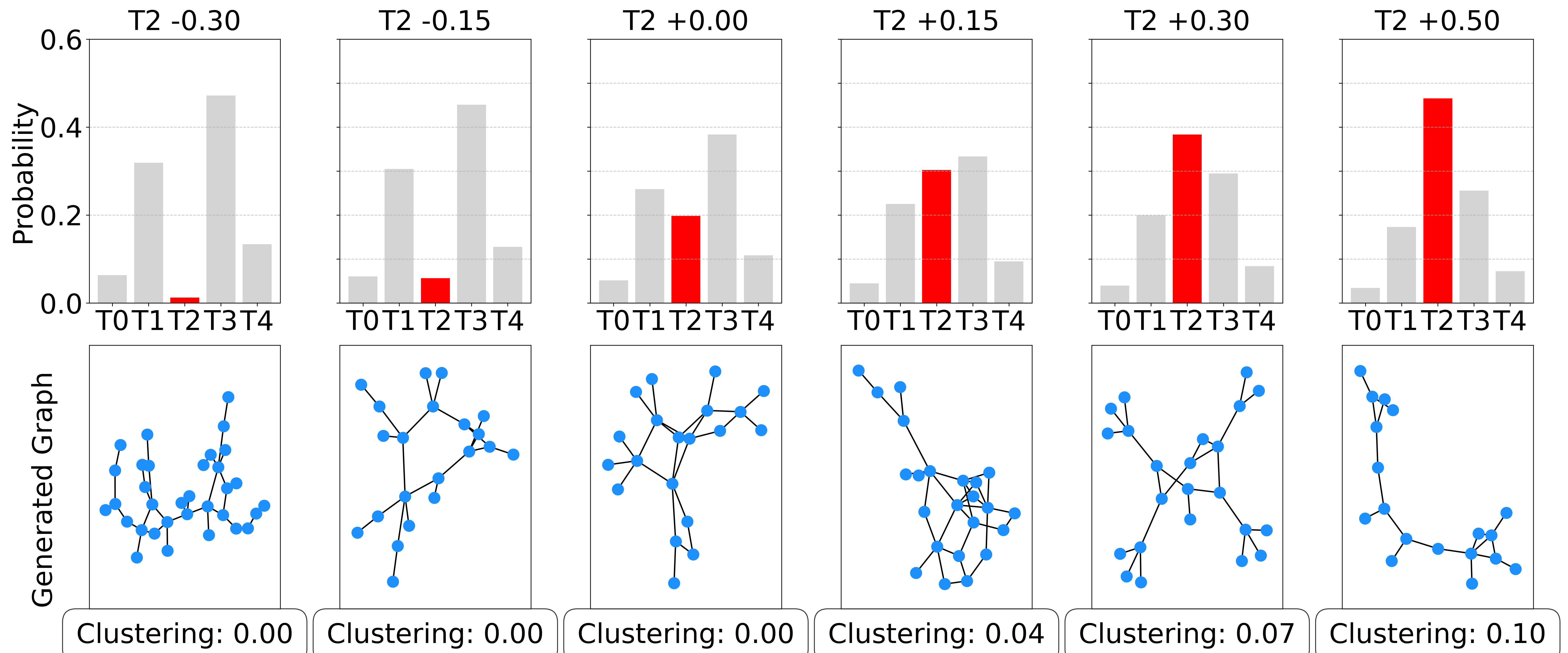}
  }
  \subfigure[T4-Clustering]{
      \includegraphics[width=0.42\textwidth]{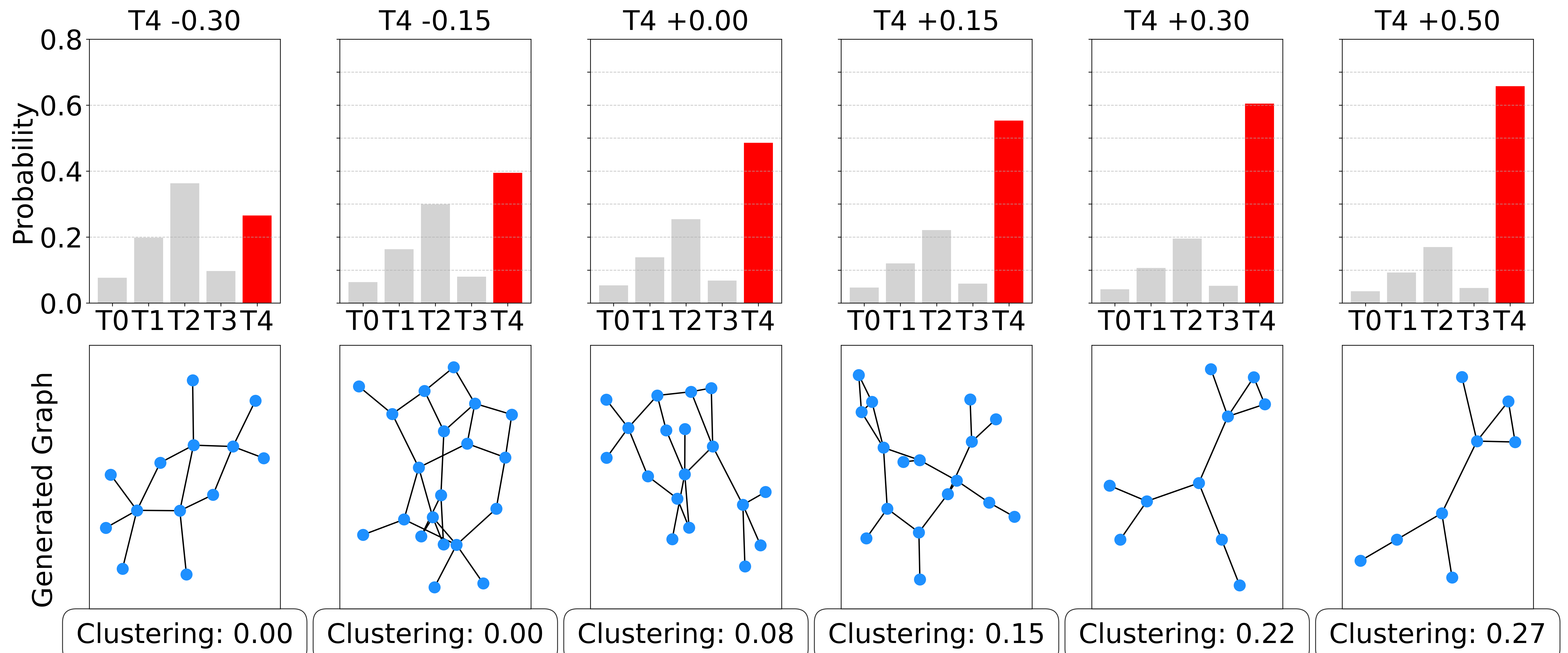}
  }
  \vspace{-10pt}
  \subfigure[T2-Density]{
      \includegraphics[width=0.42\textwidth]{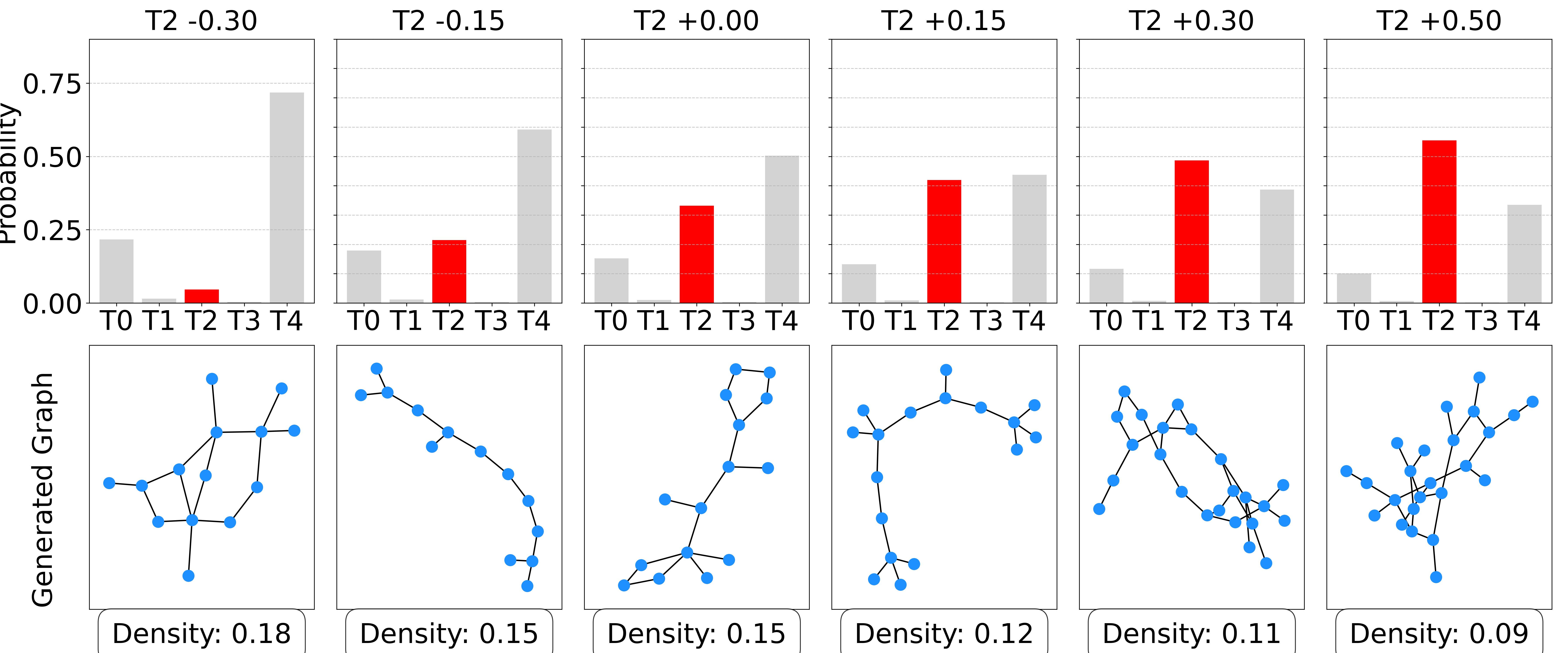}
  }
  \vspace{-10pt}
  \subfigure[T0-Diameter]{
      \includegraphics[width=0.42\textwidth]{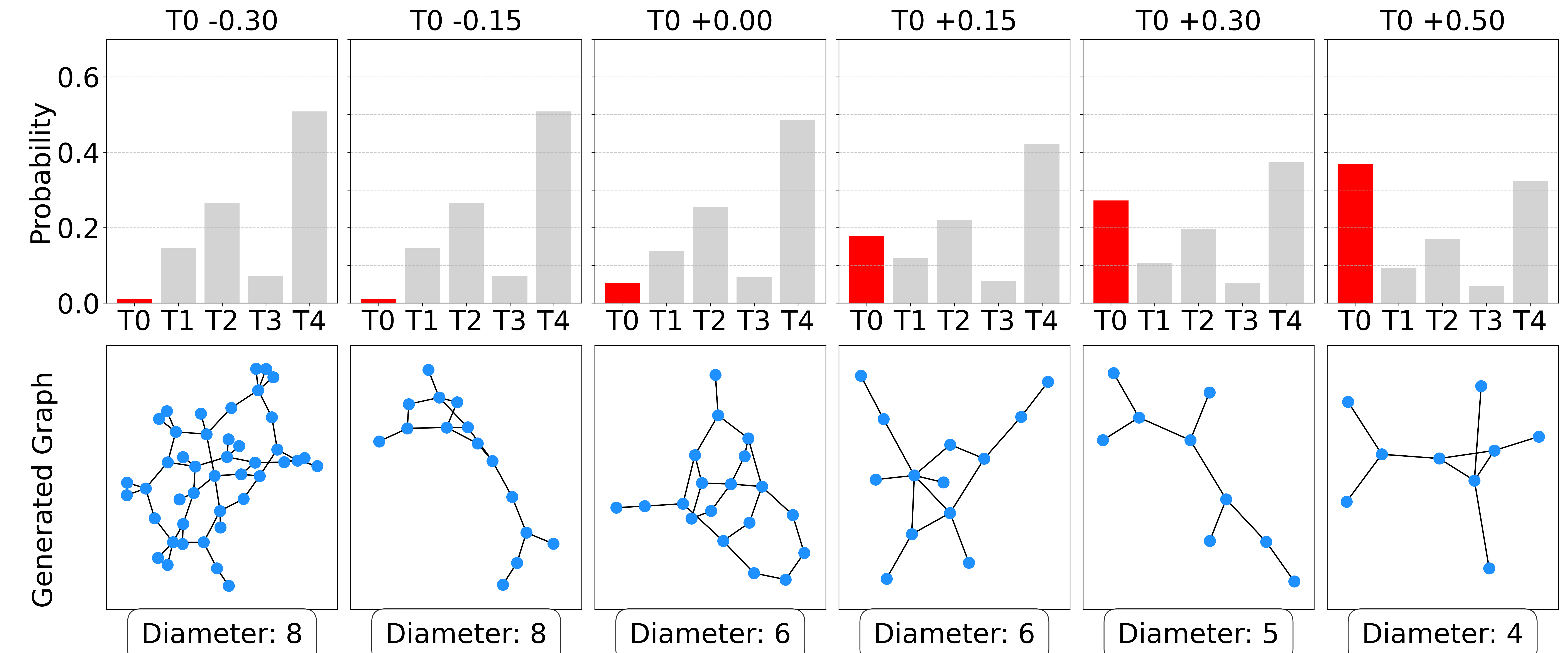}
  }
  \vspace{-10pt}
  \subfigure[T2-Node]{
      \includegraphics[width=0.42\textwidth]{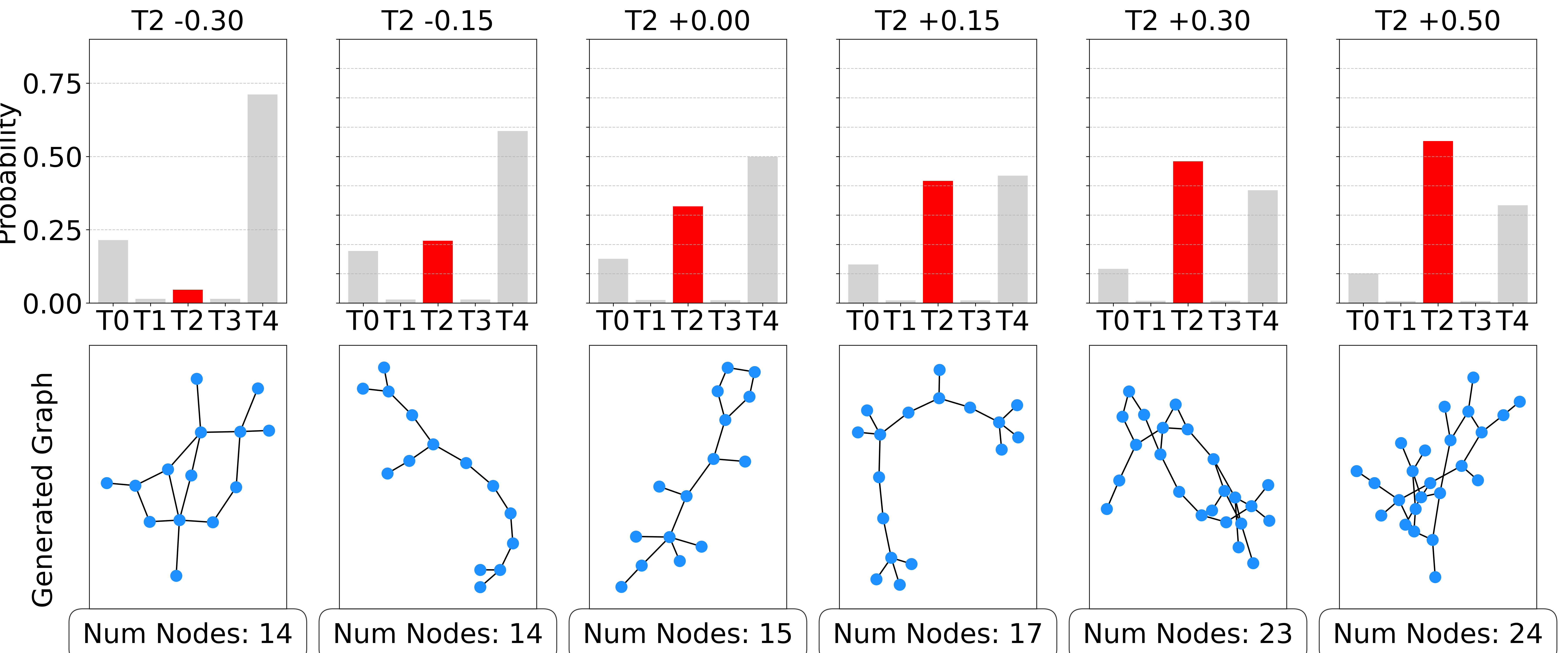}
  }
  \subfigure[T2-Square]{
      \includegraphics[width=0.42\textwidth]{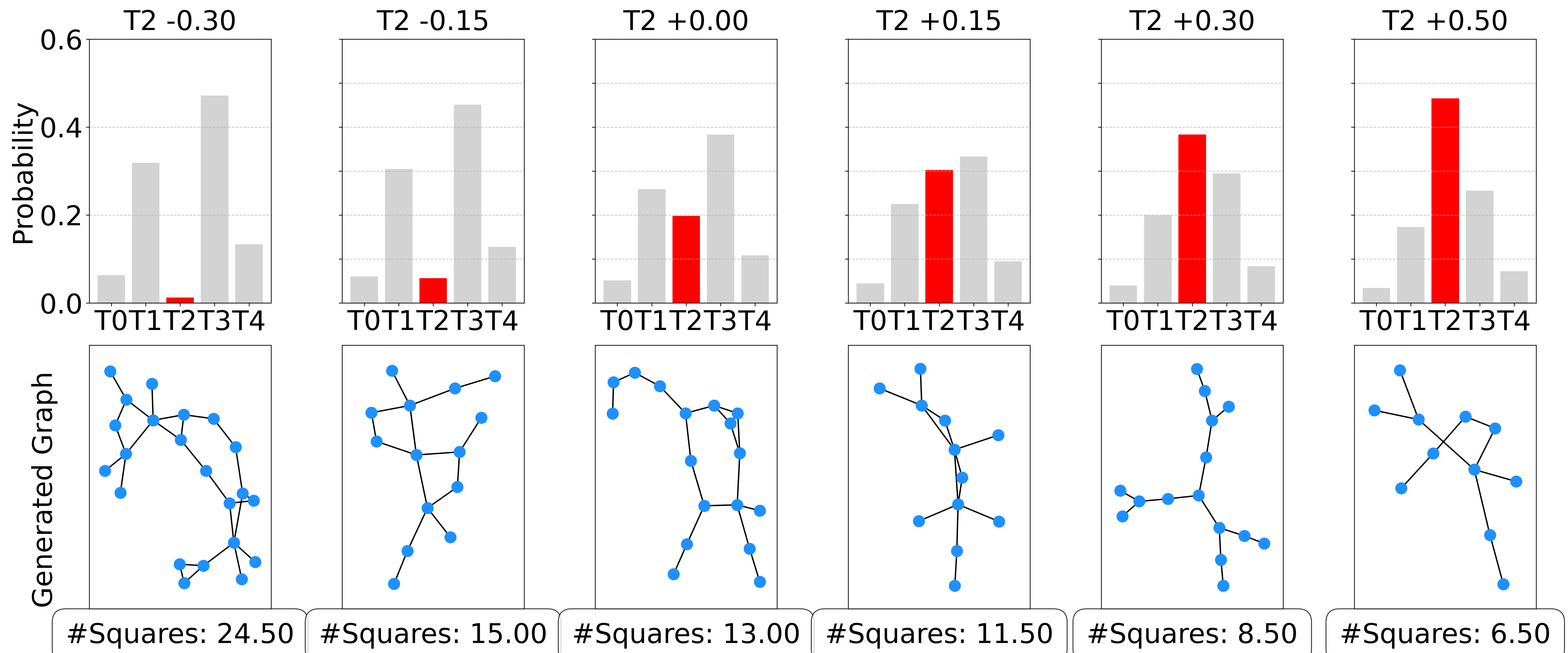}
  }
  \vspace{-10pt}
  \subfigure[T1-Square]{
    \includegraphics[width=0.42\textwidth]{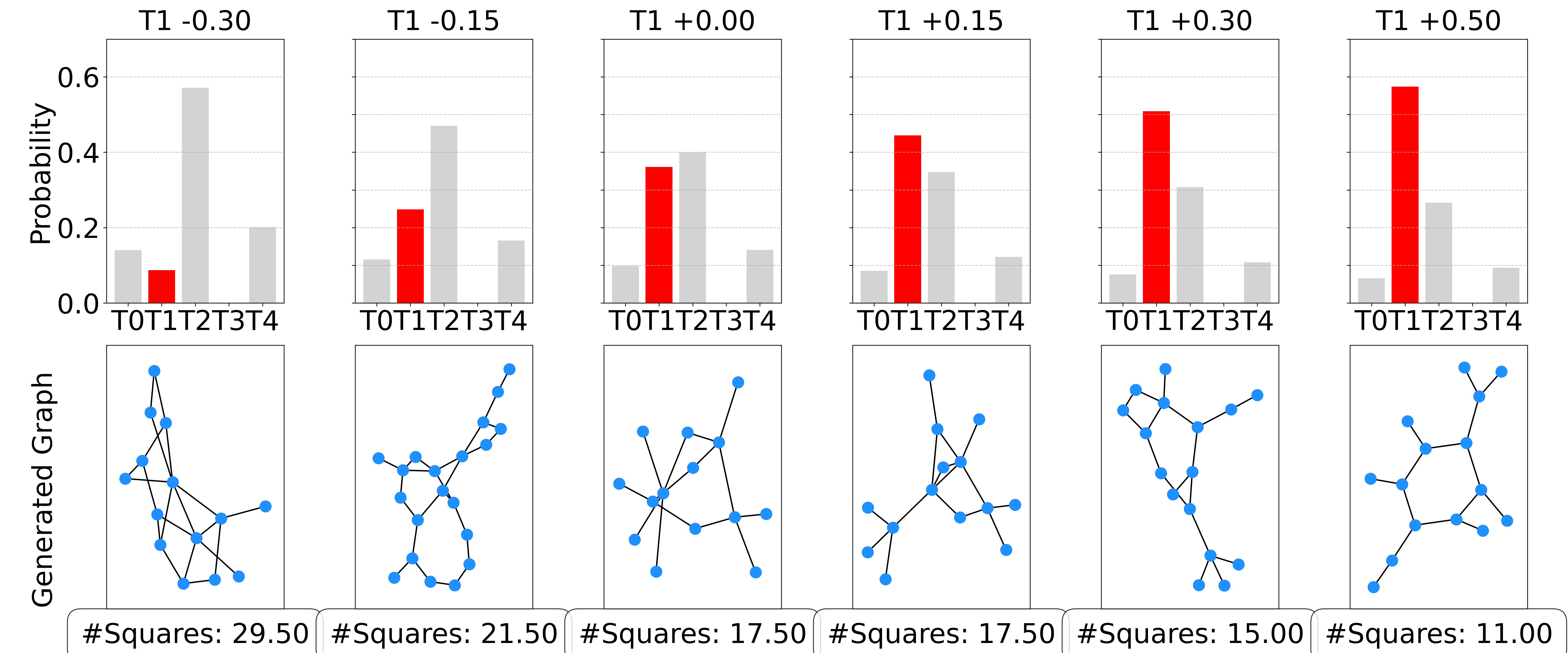}}
  \subfigure[T4-Triangle]{
      \includegraphics[width=0.42\textwidth]{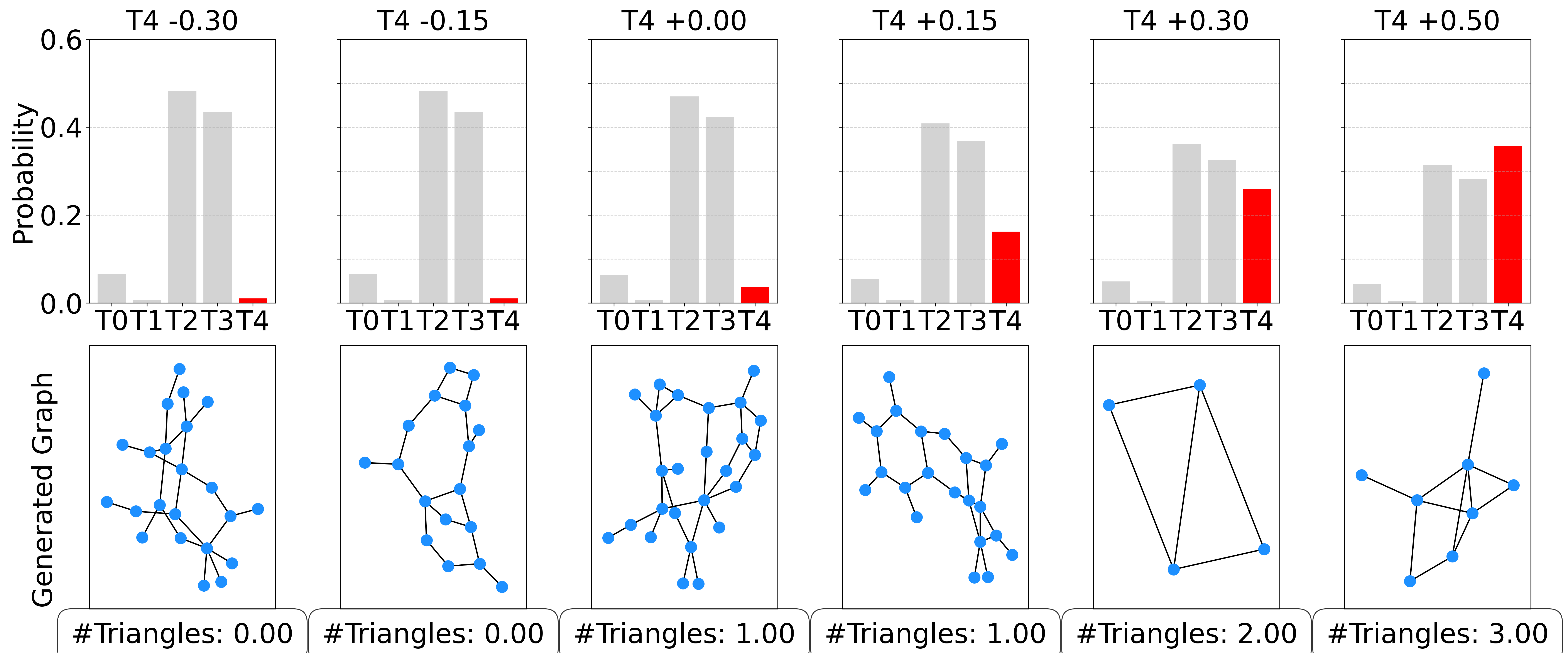}
  }
  \vspace{-10pt}
  \subfigure[T4-6cycle]{
      \includegraphics[width=0.42\textwidth]{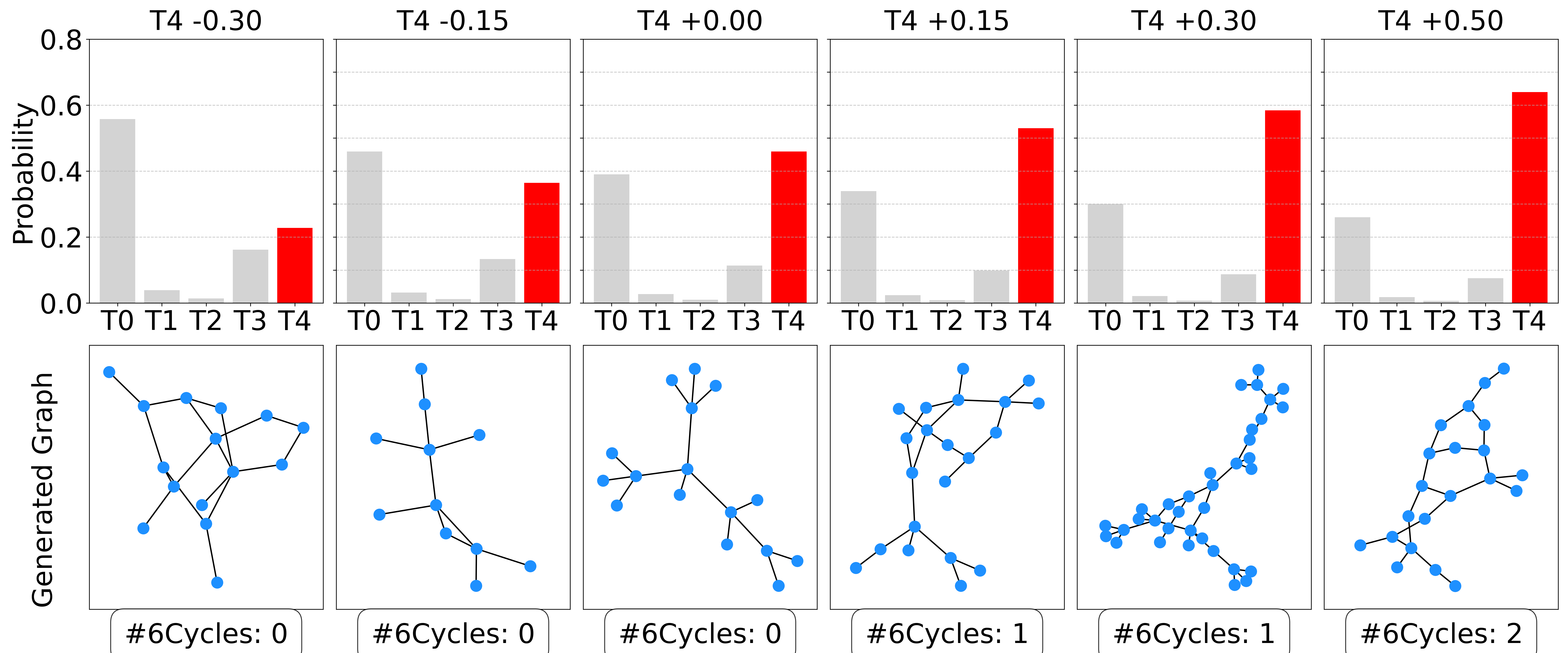}
  }
  \subfigure[T1-Modularity]{
      \includegraphics[width=0.42\textwidth]{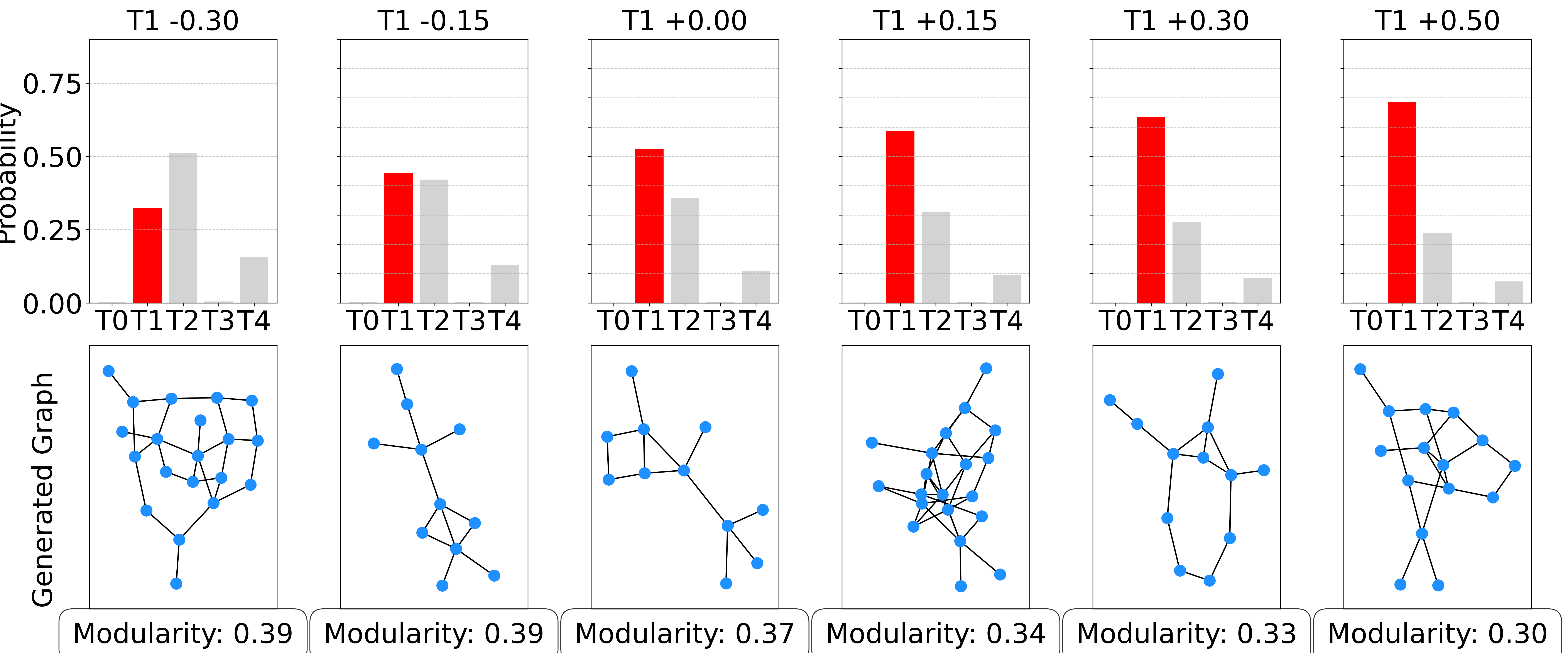}
  }
  \caption{Effect of targeted topic weight manipulation on structural properties of generated graphs. }
  \label{fig:collection}
  \end{figure}

To complement the main results, we present additional examples demonstrating how targeted topic manipulation affects various structural properties of generated graphs. Figure~\ref{fig:collection} showcases cases where adjusting individual topic weights systematically influences clustering coefficient, density, diameter, number of nodes, and the presence of specific cycle motifs (e.g., 4-cycles or squares). In each case, increasing the weight of a given topic leads to predictable and interpretable changes in the corresponding metric, reinforcing the controllability and semantic consistency of NGTM’s learned topics. These examples provide further evidence that NGTM enables fine-grained, topic-guided control over global and local graph characteristics.

\end{appendices}

\end{document}